\documentclass[10pt,twocolumn,letterpaper]{article}

\usepackage{cvpr}              

\usepackage[table,dvipsnames]{xcolor}   

\usepackage{graphicx}
\usepackage{multirow}
\usepackage{caption}
\usepackage{subcaption}
\usepackage{wrapfig}
\usepackage{tikz}
\usepackage{stfloats}  
\usepackage{etoolbox}
\AfterEndEnvironment{strip}{\leavevmode}
\usepackage[section]{placeins}
\usepackage{float}
\usepackage{adjustbox}

\definecolor{cvprblue}{rgb}{0.21,0.49,0.74}
\usepackage[pagebackref,breaklinks,colorlinks,citecolor=cvprblue]{hyperref}

\def\paperID{322} 
\def\confYear{2026\xspace}

\title{A Streamlined Attention-Based Network for Descriptor Extraction}

\author{
    Mattia D'Urso$^1$
    \quad
    Emanuele Santellani$^1$
    \quad
    Christian Sormann$^2$
    \quad
    Mattia Rossi$^2$ \\
    \quad
    Andreas Kuhn$^2$
    \quad
    Friedrich Fraundorfer$^1$\\
    {\normalsize 
        $^1$Graz University of Technology \{name.surname@tugraz.at\},
        $^2$Sony  \{name.surname@sony.com\}
    }
}

\begin{document}

\begin{figure*}
    \renewcommand\twocolumn[1][]{#1}%
    \maketitle
    \centering
    \begin{subfigure}[c]{0.40\textwidth}
        \includegraphics[width=\linewidth]{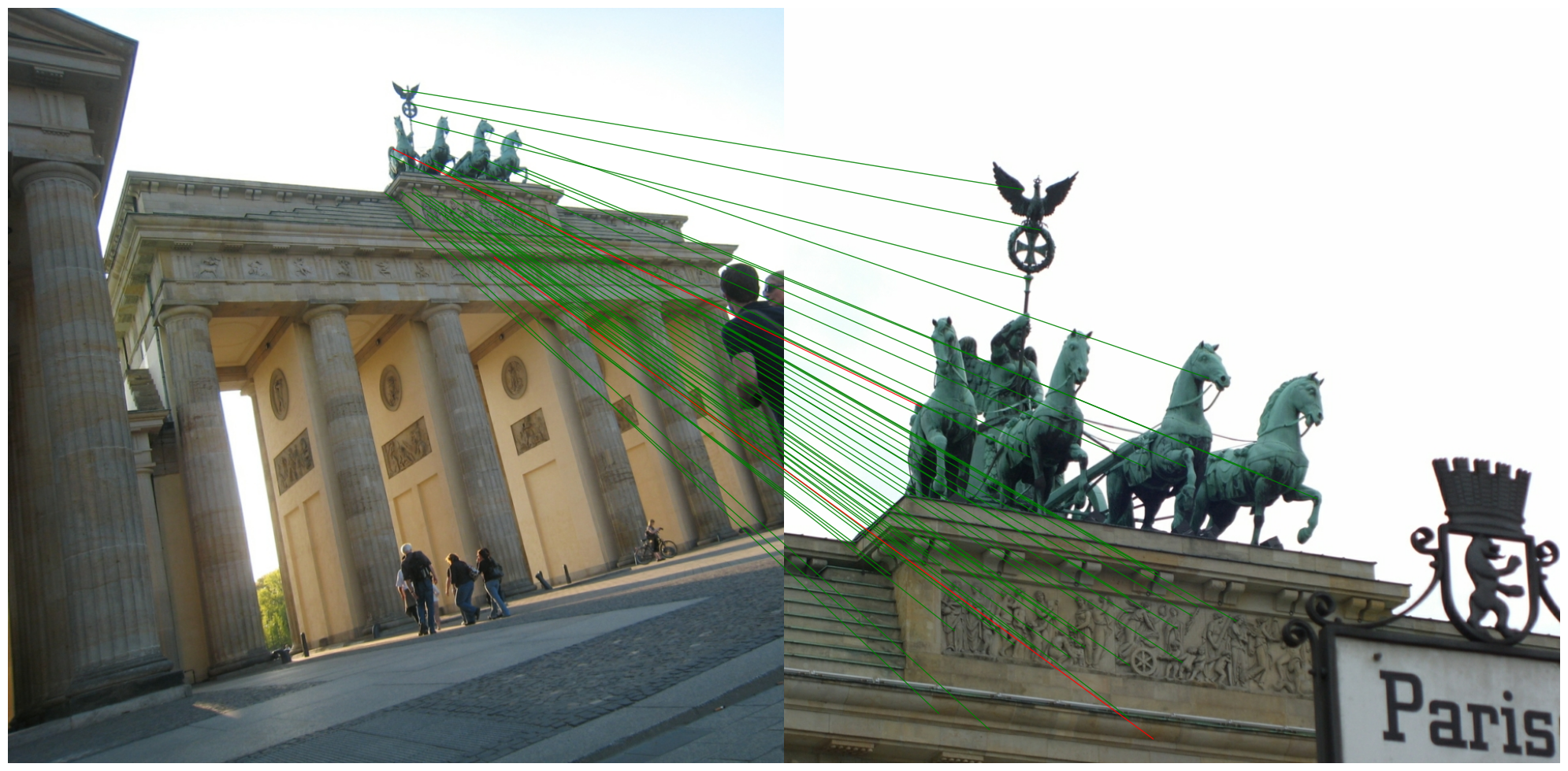}
        \caption{ALIKED with \textup{SAN\textsc{Desc}} descriptors.}
        \label{fig:qual-a}
    \end{subfigure}
    \hfill
    \begin{subfigure}[c]{0.585\textwidth}
        \includegraphics[width=\linewidth]{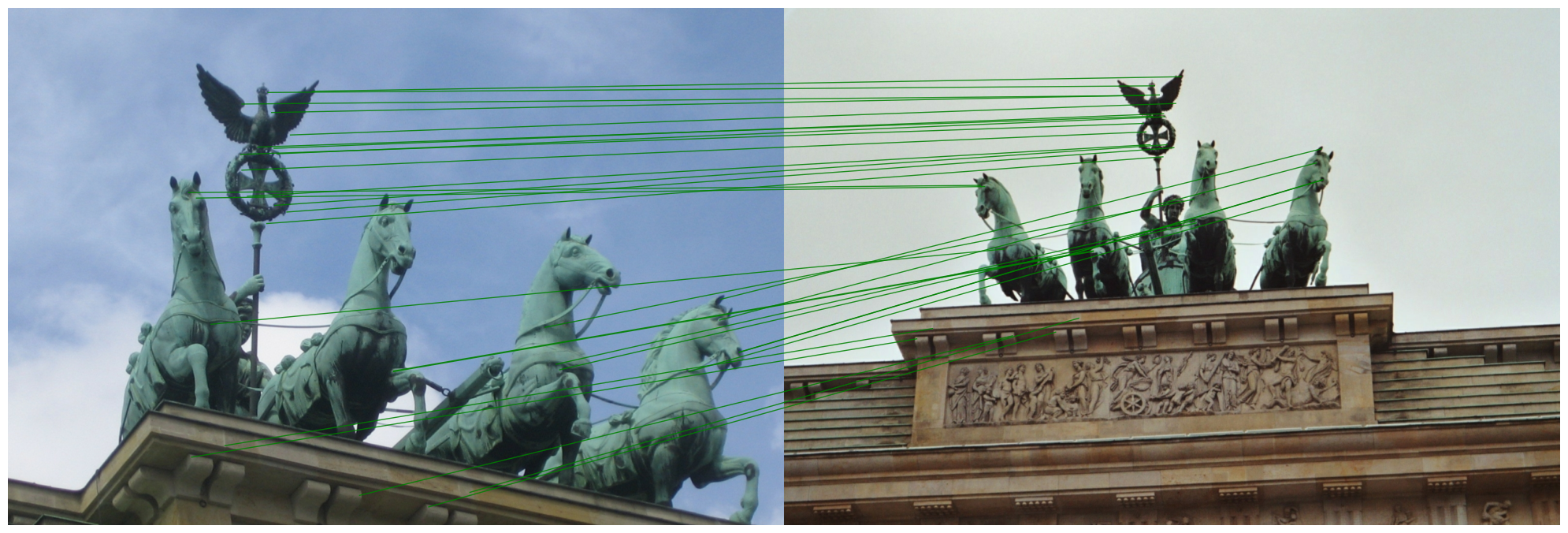}
        \caption{DeDoDe with \textup{SAN\textsc{Desc}} descriptors.}
        \label{fig:qual-b}
    \end{subfigure}

    \vspace{1mm}

    \begin{subfigure}[c]{0.40\textwidth}
        \includegraphics[width=\linewidth]{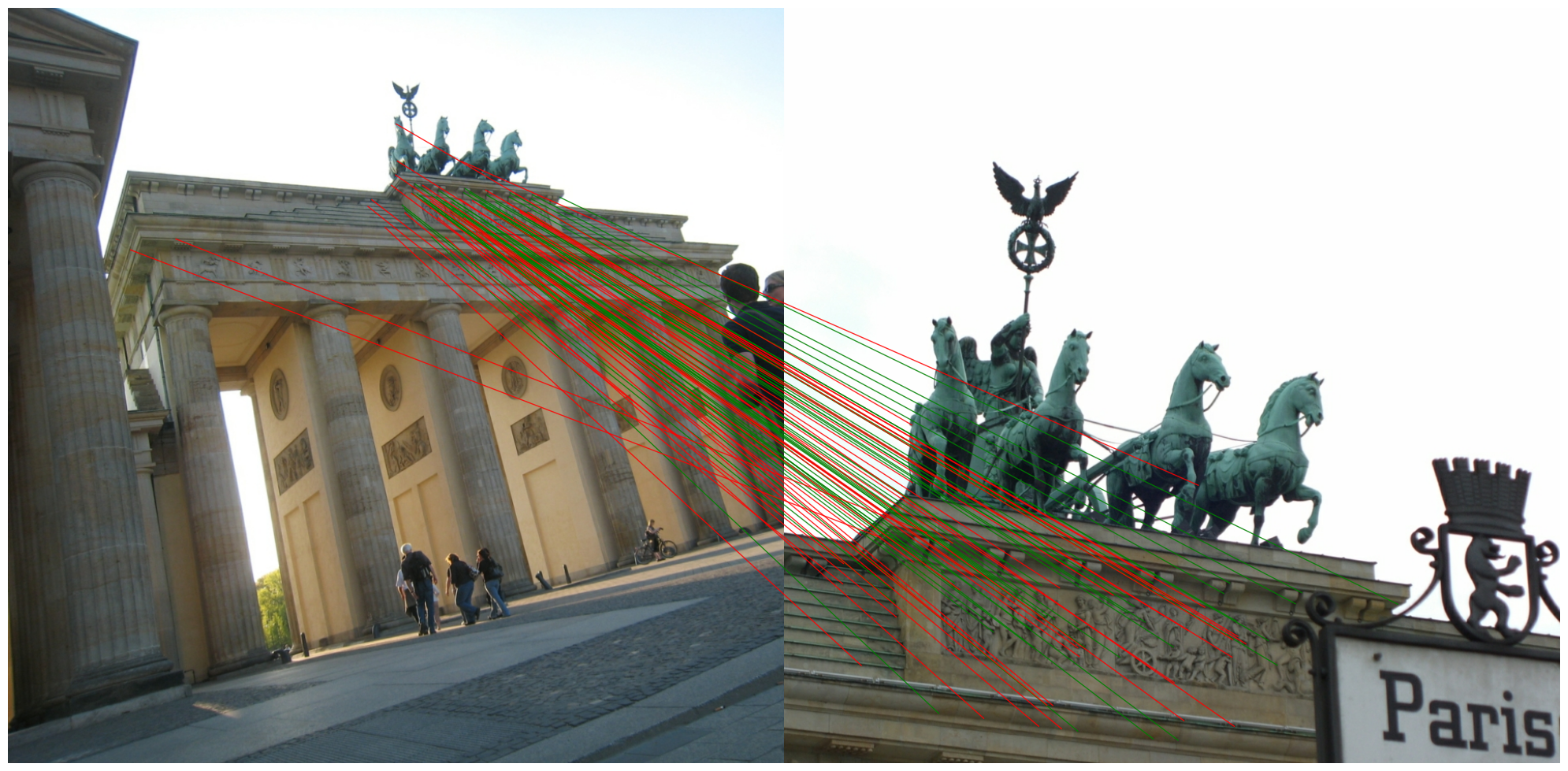}
        \caption{ALIKED with original descriptors.}
        \label{fig:qual-c}
    \end{subfigure}
    \hfill
    \begin{subfigure}[c]{0.585\textwidth}
        \includegraphics[width=\linewidth]{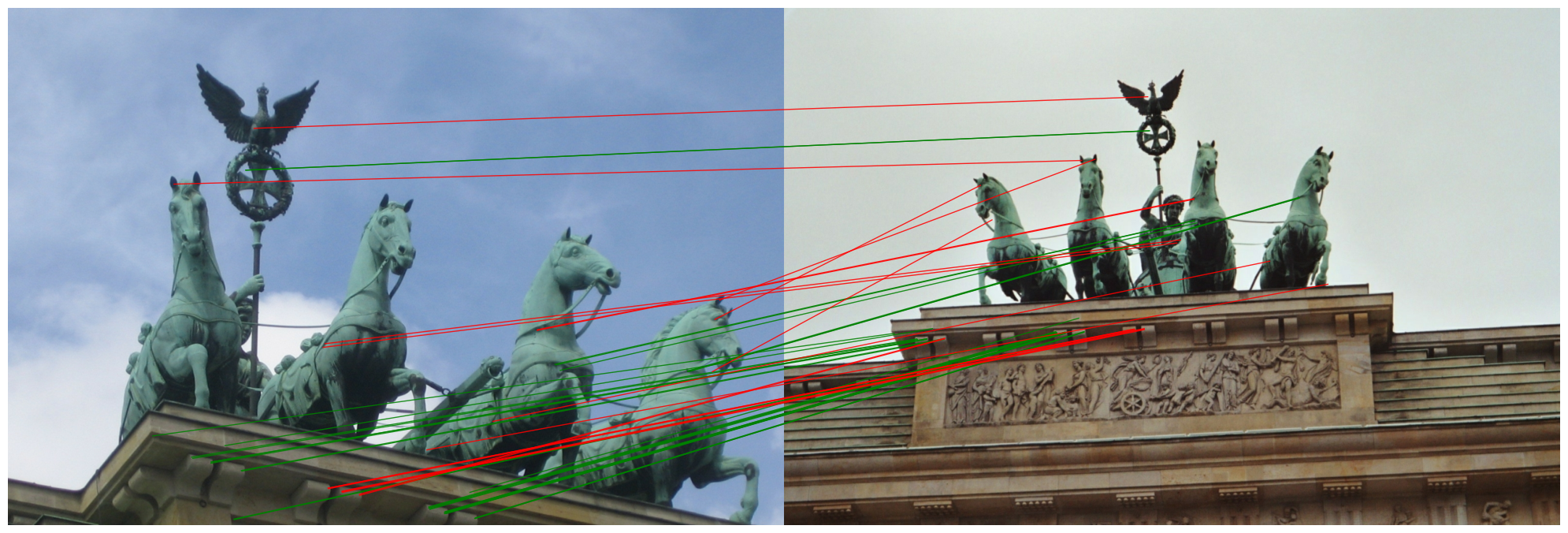}
        \caption{DeDoDe with original descriptors.}
        \label{fig:qual-d}
    \end{subfigure}

    \caption{\textbf{Qualitative comparison.} Examples of local feature matching between pairs of images exhibiting large scale differences. Inliers are displayed in green; outliers in red.}
    \label{fig:teaser}
\end{figure*}

\begin{abstract}

We introduce \textbf{\textup{SAN\textsc{Desc}}} a \underline{S}treamlined \underline{A}ttention-based \underline{N}etwork for \underline{Desc}riptor extraction that aims to improve on existing architectures for keypoint description.

Our descriptor network learns to compute descriptors that improve matching without modifying the underlying keypoint detector.
We employ a revised \mbox{U-Net-like} architecture enhanced with Convolutional Block Attention Modules and residual paths, enabling effective local representation while maintaining computational efficiency.
We refer to the building blocks of our model as Residual U-Net Blocks with Attention. The model is trained using a modified triplet loss in combination with a curriculum learning–inspired hard negative mining strategy, which improves training stability.

Extensive experiments on HPatches, MegaDepth-1500, and the Image Matching Challenge 2021 show that training \textup{SAN\textsc{Desc}} on top of existing keypoint detectors leads to improved results on multiple matching tasks compared to the original keypoint descriptors. At the same time, \textup{SAN\textsc{Desc}} has a model complexity of just 2.4 million parameters.

As a further contribution, we introduce a new urban dataset featuring 4K images and pre-calibrated intrinsics, designed to evaluate feature extractors. On this benchmark, \textup{SAN\textsc{Desc}} achieves substantial performance gains over the existing descriptors while operating with limited computational resources.

\end{abstract}

\vspace{-5pt}
\section{Introduction}
\label{sec:intro}

\begin{figure*}
  \centering

  \begin{subfigure}[c]{0.60\textwidth}
    \includegraphics[width=\linewidth]{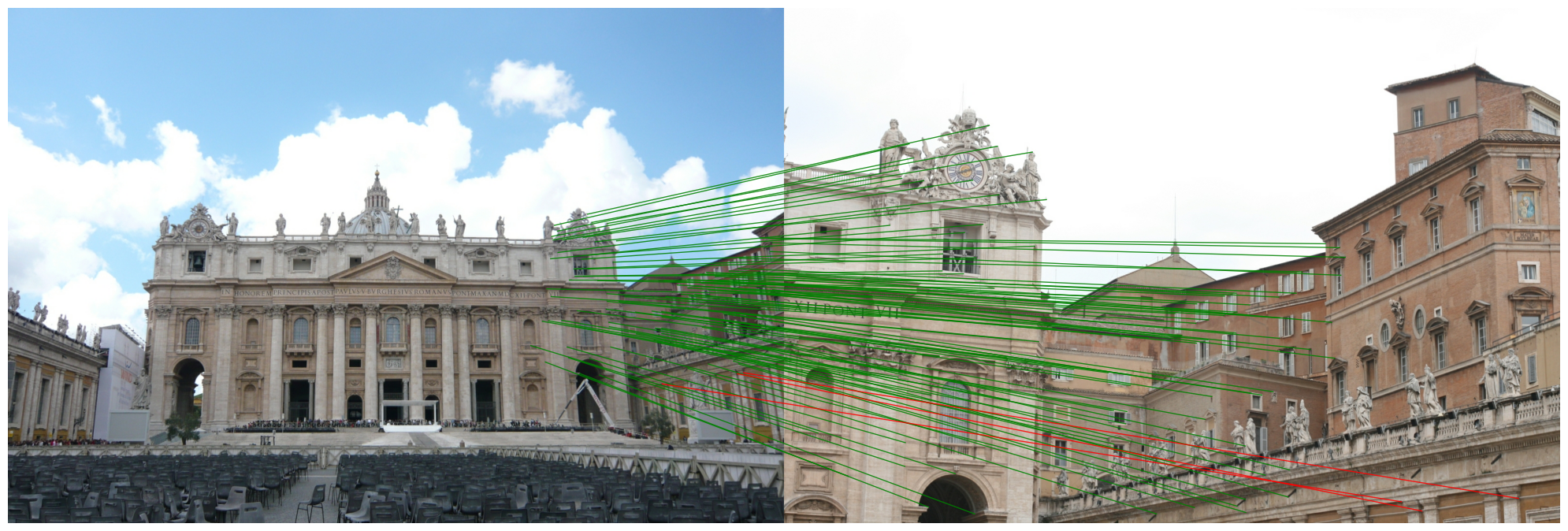}
    \caption{SuperPoint with \textup{SAN\textsc{Desc}} descriptors.}
    \label{fig:teaser2-a}
  \end{subfigure}
  \hfill
  \begin{subfigure}[c]{0.385\textwidth}
    \includegraphics[width=\linewidth]{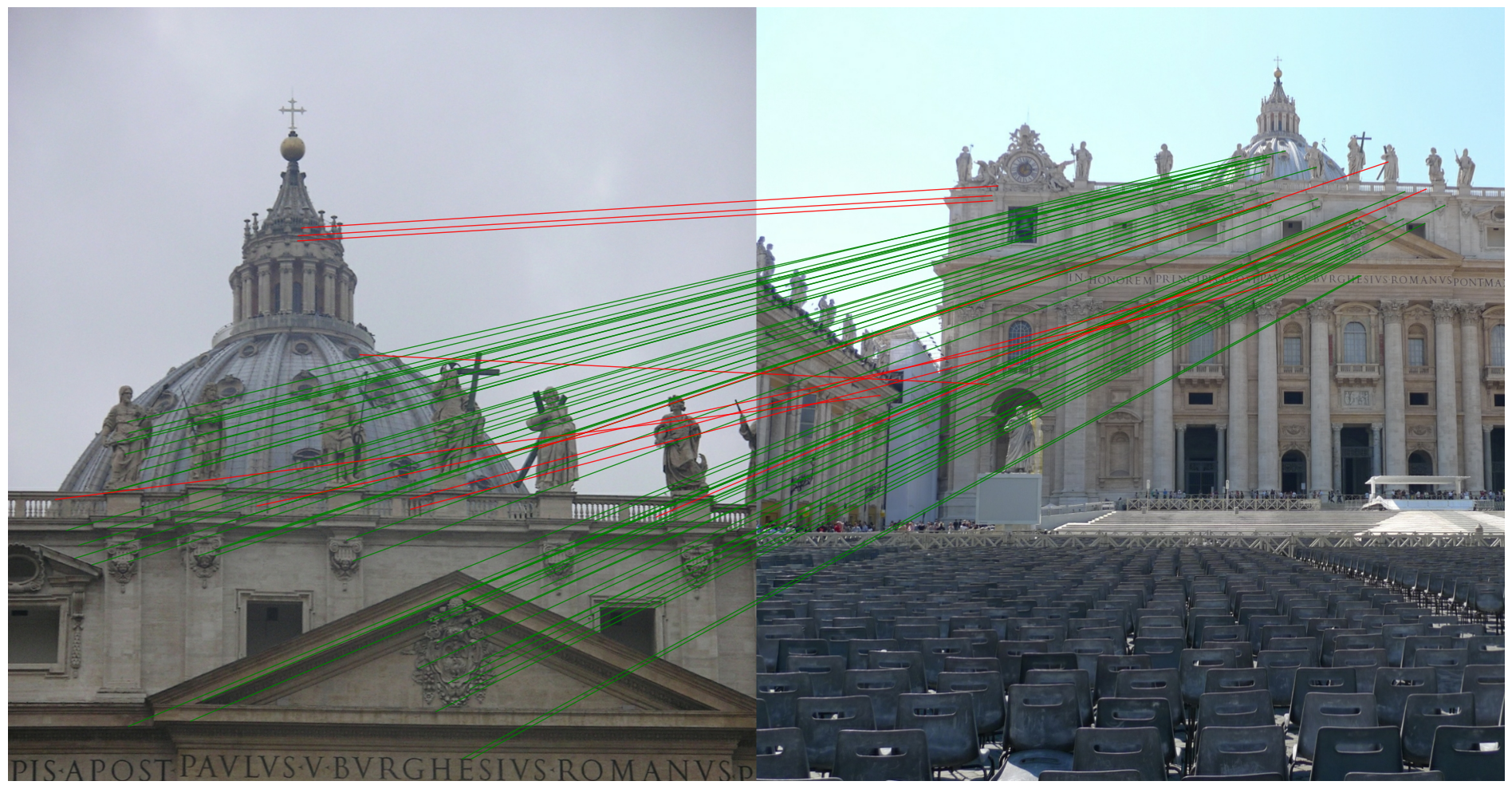}
    \caption{DISK with \textup{SAN\textsc{Desc}} descriptors.}
    \label{fig:teaser2-b}
  \end{subfigure}

  \vspace{1mm}

  \begin{subfigure}[c]{0.60\textwidth}
    \includegraphics[width=\linewidth]{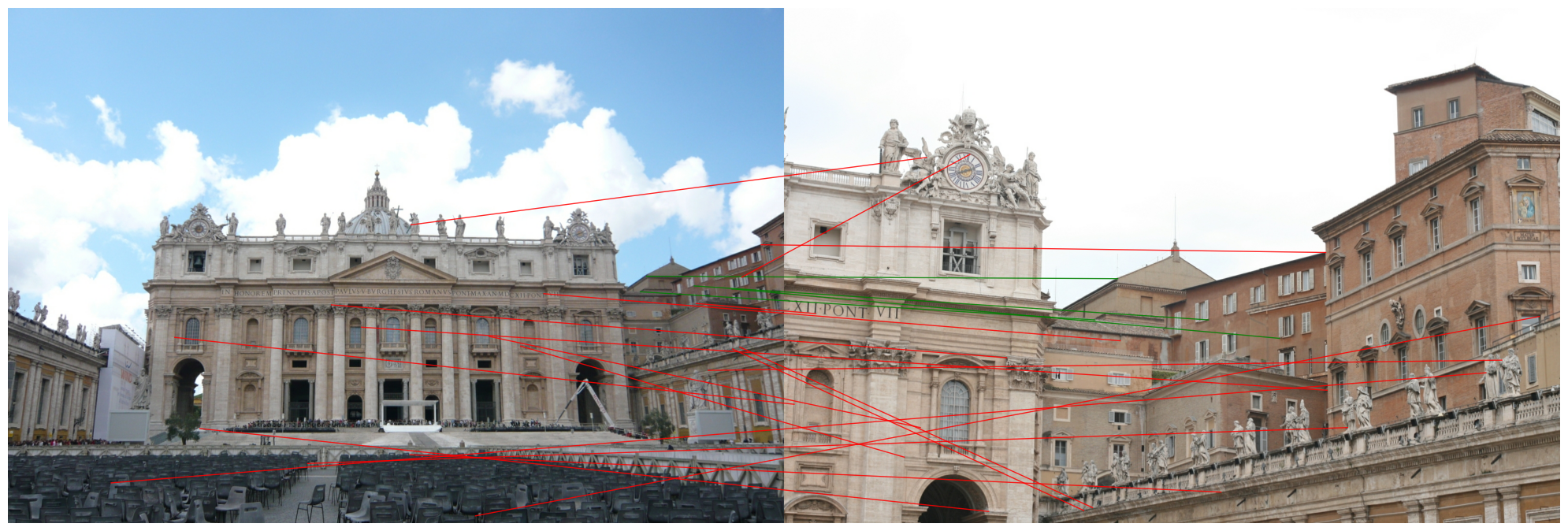}
    \caption{SuperPoint with original descriptors.}
    \label{fig:teaser2-c}
  \end{subfigure}
  \hfill
  \begin{subfigure}[c]{0.385\textwidth}
    \includegraphics[width=\linewidth]{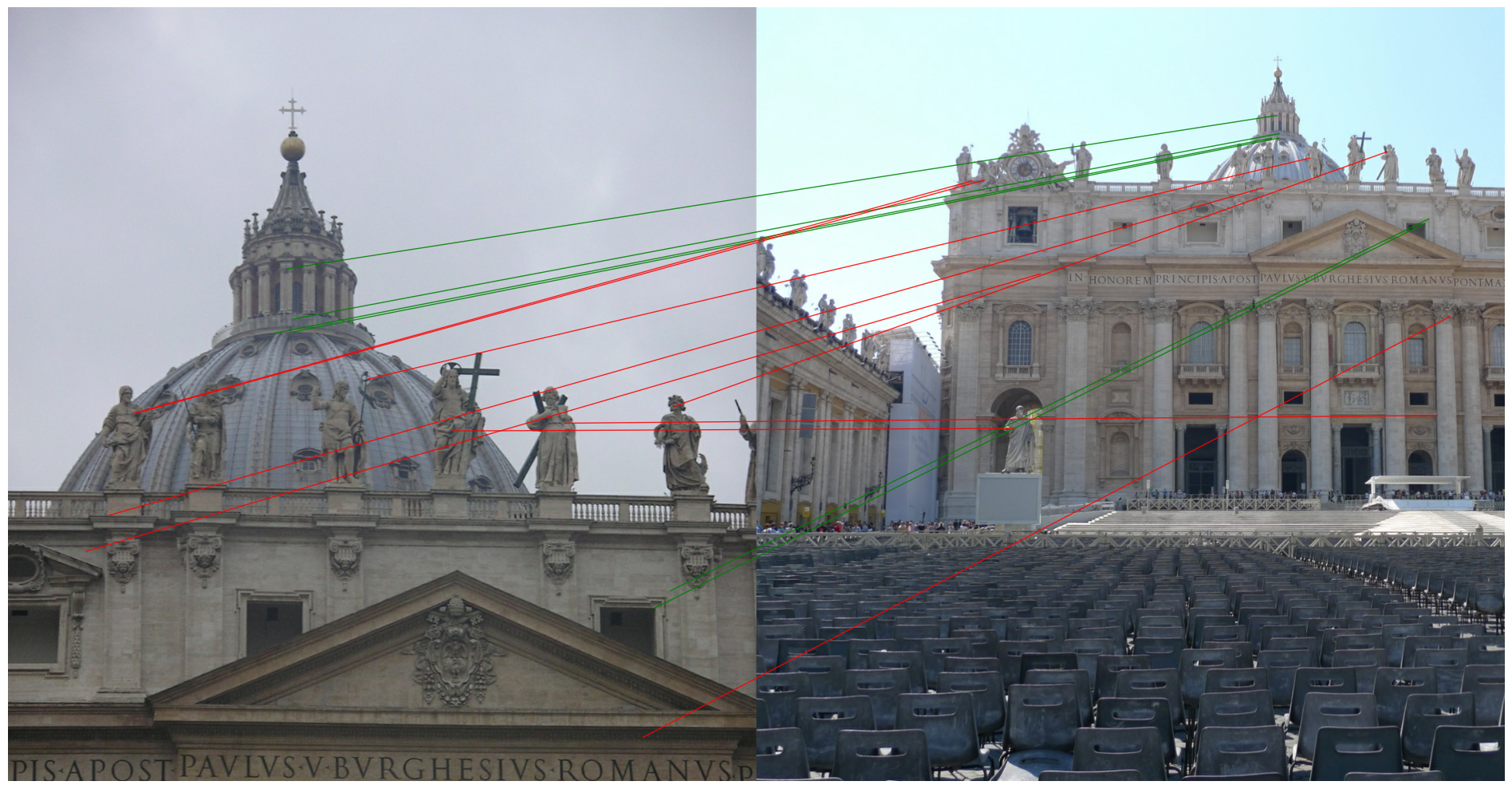}
    \caption{DISK with original descriptors.}
    \label{fig:teaser2-d}
  \end{subfigure}
  \caption{\textbf{Qualitative comparison.} Examples of local feature matching between pairs of images exhibiting large scale differences. Inliers are displayed in green; outliers in red.}
  \label{fig:teaser2}
\end{figure*}

Finding reliable correspondences between images is a fundamental problem in computer vision, with key applications in Structure-from-Motion (SfM) \cite{schoenberger2016sfm}, Visual Localization (VL) \cite{sattler2018benchmarking} and Object Tracking \cite{zhou2009object}. The classical SfM pipeline involves three main stages: \mbox{feature} extraction (keypoint detection and description), feature matching and geometric verification, and 3D mapping. While early methods, such as SIFT \cite{sift} and SURF \cite{bay2008surf}, use handcrafted features, modern descriptor learning techniques, such as L2-Net \cite{tian2017l2net} and HardNet \cite{mishchuk2017working}, have significantly improved matching performance.

A major shift occurred with the introduction of all-in-one networks, such as SuperPoint \cite{detone2018superpoint}, which jointly optimize keypoint detection and description within a single unified architecture. However, coupling detection and description introduces interdependencies that can constrain flexibility and degrade overall performance, while recent works indicate that decoupling the two tasks not only simplifies the training process but also enhances robustness and generalization capabilities~\cite{li2022decoupling}. 
A growing line of methods, including XFeat~\cite{potje2024xfeat}, S-TREK~\cite{santellani2023strek}, and DeDoDe~\cite{edstedt2024dedode} further highlight the benefits of decoupling keypoint from descriptor learning.

Despite the strong accuracy of DeDoDe’s decoupled design~\cite{edstedt2024dedode}, its descriptors' compute and memory demand is a bottleneck in long-duration sequences, large-scale SfM reconstructions, and with high-resolution imagery where more pixels inflate keypoints/descriptors and increase the cost of extraction and matching. This overhead limits the deployment on resource-constrained platforms and slows pipelines processing millions of descriptors, underscoring the need for a scalable descriptor architecture that preserves robustness while substantially reducing training and inference costs. 

In order to address these issues, we propose \textup{SAN\textsc{Desc}}, a lightweight descriptor network architecture that aims at providing existing keypoint detectors with more discriminative descriptors than their native ones. In particular, \textup{SAN\textsc{Desc}} leverages a compact U-Net-like architecture enhanced with an attention mechanism from the Convolutional Block Attention Module (CBAM) \cite{woo2018cbam}, which allows for effective local feature representation while maintaining computational efficiency. The model is trained using a modified triplet-based loss combined with a curriculum-inspired hard negative mining strategy, improving both stability and discriminative power. \textup{SAN\textsc{Desc}} leverages the sequential dependence of description on detection: its descriptor performs best when trained on the keypoint distribution of the target detector, as discussed in Section~\ref{sec:discussion}.

Extensive evaluations on HPatches, MegaDepth-1500, and the Image Matching Challenge 2021 show that \textup{SAN\textsc{Desc}} consistently provides more robust descriptors across diverse detectors. Moreover, to evaluate performance on high-resolution imagery, we introduce a new high-resolution urban dataset with pre-calibrated intrinsics and poses estimated via COLMAP. This benchmark not only tests a model’s ability to extract reliable local features in high-resolution images but also stresses its efficiency under constrained computational budgets. 

Figures \ref{fig:teaser}, \ref{fig:teaser2} and \ref{fig:teaser3} compare matches obtained with \textup{SAN\textsc{Desc}} against other method’s native descriptors across challenging cases. 
These figures highlight failure cases and limitations of state-of-the-art methods. They further demonstrate that our proposed descriptors can address these challenging scenarios. Additional examples are provided in the Supplementary Material.

Our main contributions are as follows: 
\begin{itemize}

    \item We introduce a new attention-based descriptor architecture with only 2.4M parameters, trained using a revised triplet loss combined with a curriculum learning–inspired hard negative mining approach for stable and efficient learning. 
    
    \item We present a new real-world benchmark consisting of carefully collected, high-resolution urban scenes captured with pre-calibrated intrinsics, extending the typical evaluation scheme with a practically relevant scenario. 
    
    \item With extensive experiments across several established benchmarks, we demonstrate significant image matching improvements using the proposed \textup{SAN\textsc{Desc}} descriptor.

\end{itemize}

\begin{figure*}
  \centering

  \begin{subfigure}[c]{0.49\textwidth}
    \includegraphics[width=\linewidth]{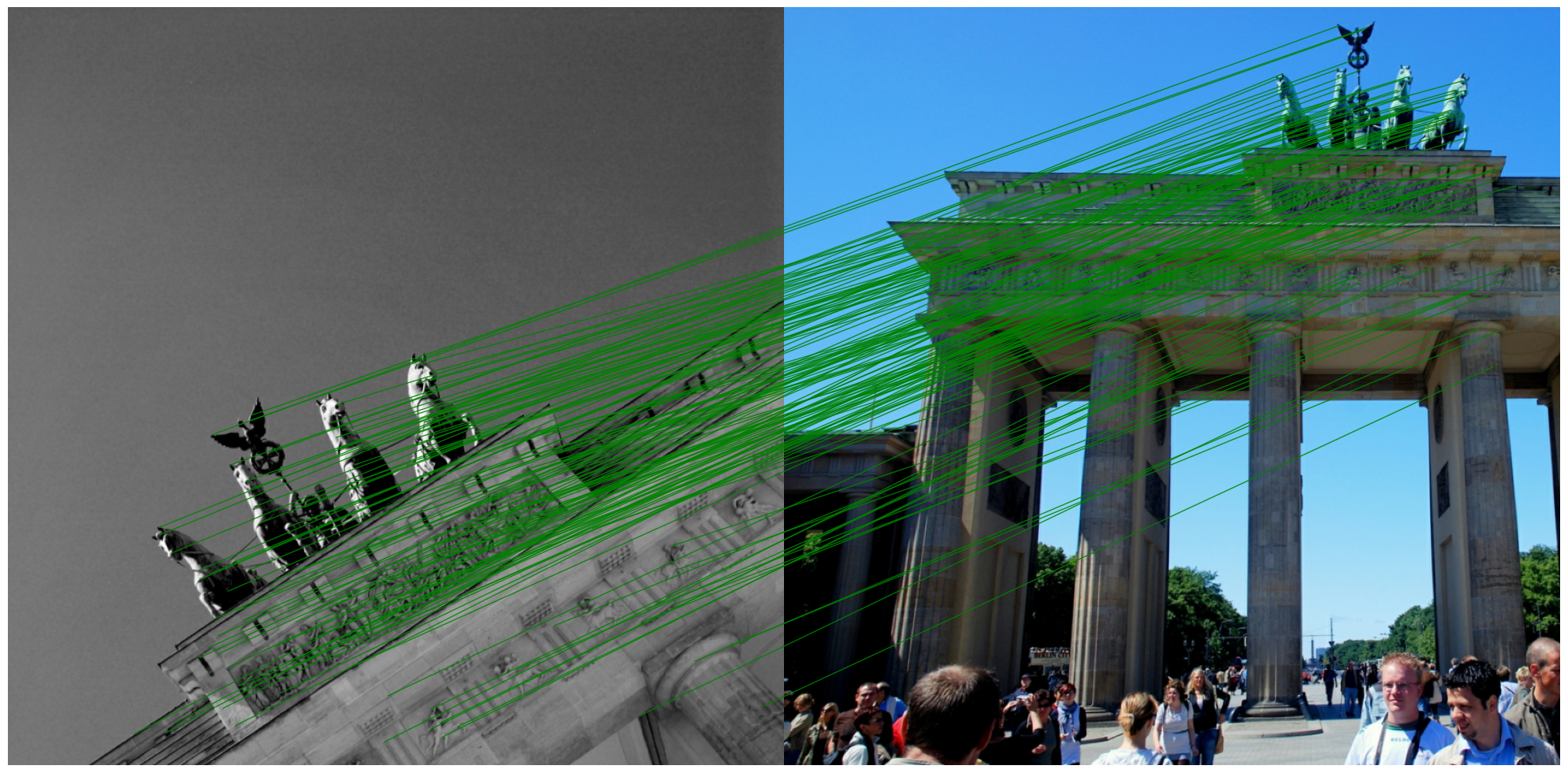}
    \caption{RIPE with \textup{SAN\textsc{Desc}} descriptors.}
    \label{fig:teaser3-a}
  \end{subfigure}
  \hfill
  \begin{subfigure}[c]{0.49\textwidth}
    \includegraphics[width=\linewidth]{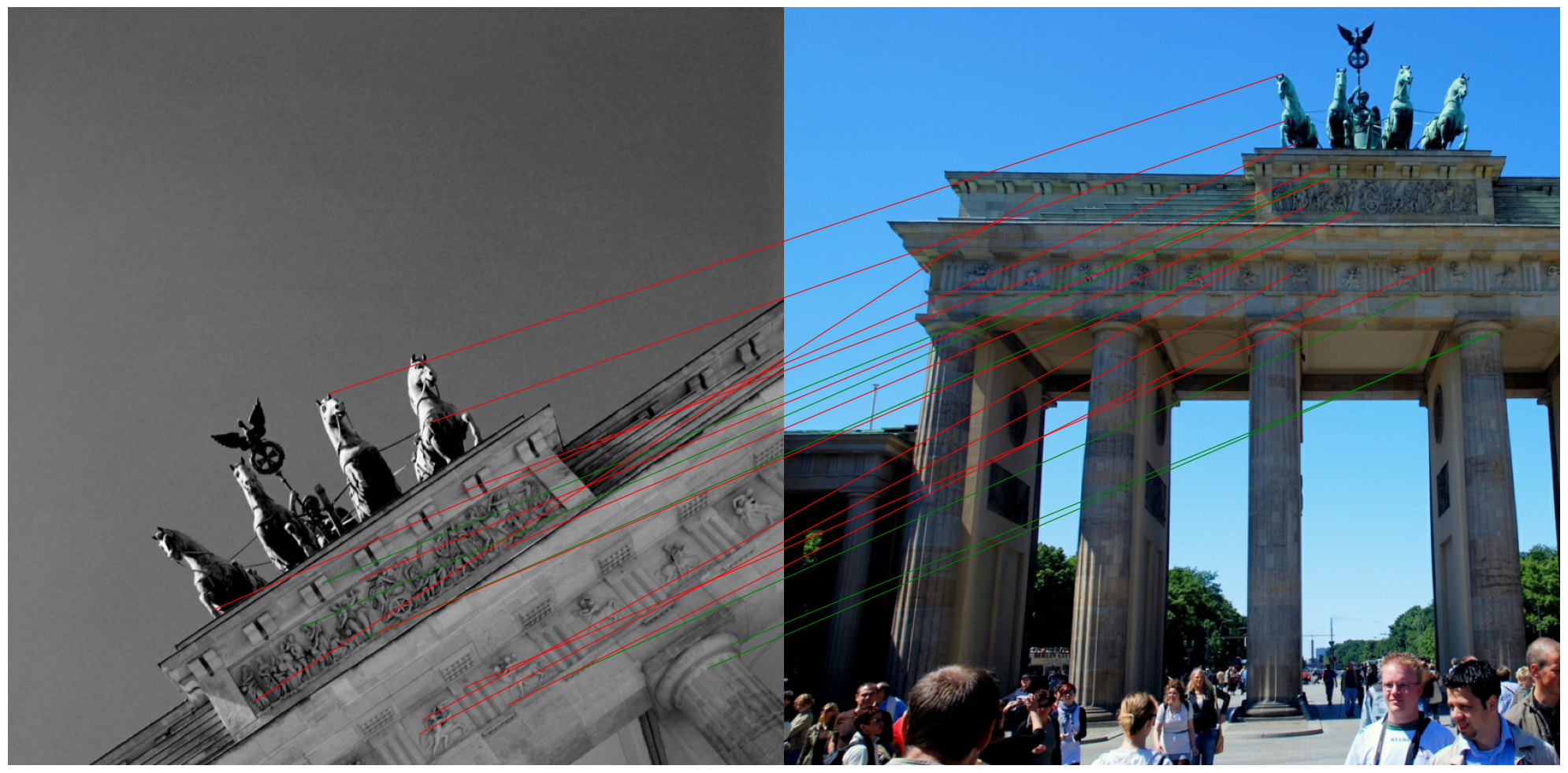}
    \caption{RIPE with original descriptors.}
    \label{fig:teaser3-b}
  \end{subfigure}

  \caption{\textbf{Qualitative comparison.} Example of local feature matching between a 
  pair of images exhibiting large scale differences. Inliers are displayed in green; outliers in red.}
  \label{fig:teaser3}
  \vspace{-6pt}
\end{figure*}

\section{Related Work}
\label{sec:related}

\paragraph{Learned Patch Descriptors.}
In recent years, a significant number of works in the literature have studied the advantages of deep learning-based approaches compared to traditional handcrafted local descriptors such as SIFT~\cite{sift} and SURF~\cite{bay2008surf}.
Early works, such as MatchNet~\cite{han2015matchnet} and DeepDesc~\cite{simo2015dldcf}, focused on learning descriptors from image patches extracted at given keypoint locations (often SIFT detections). These initial methods demonstrated the advantage of training deep networks to compare local patches, significantly improving upon classical descriptors. Subsequent works refined descriptor learning through improved objectives and mining strategies. For example, the triplet ranking loss~\cite{balntas2016triplet} became a standard for training robust descriptors by maximizing the margin between the hardest positive and negative pairs in a batch, leading to methods such as L2-Net~\cite{tian2017l2net} and HardNet~\cite{mishchuk2017working}. Additional regularization techniques, such as second-order similarity in SOSNet~\cite{tian2019sosnet}, boosted descriptor invariance. By the end of the 2010s, learned patch descriptors achieved superior matching accuracy and robustness compared to handcrafted ones, as demonstrated in benchmarks such as HPatches~\cite{balntas2017hpatches} and the Image Matching Challenge 2021~\cite{jin2021image}.

\paragraph{Joint Detection and Description.}
Rather than relying on an external detector, a second generation of methods learned to perform detection and description jointly within a single network. SuperPoint~\cite{detone2018superpoint} pioneered self-supervised joint learning in this context, inspiring a series of end-to-end models such as D2-Net~\cite{dusmanu2019d2net}, R2D2~\cite{revaud2019r2d2}, and MD-Net~\cite{santellani2022mdnet}. Notably, DISK~\cite{tyszkiewicz2020disk} integrated a reinforcement learning approach optimizing a probabilistic detector-descriptor network with a cycle-consistency reward, allowing back-propagation through the detection process and achieving state-of-the-art results on the Image Matching Challenge 2021 \cite{jin2021image}. 

\paragraph{The Rationale for Decoupling.}
An emerging line of research revisits the idea of decoupling detection and description, based on the observation that training keypoint detectors and descriptors together can create harmful interdependencies that reduce overall performance~\cite{li2022decoupling}. Li et al.~\cite{li2022decoupling} showed that training the detector independently, without forcing it to align with the descriptor during training, produces higher-quality keypoints, since the detector can optimize solely for its own task rather than compromising for descriptor compatibility. 

\paragraph{Modern Decoupled Architectures.}
Edstedt et al.~\cite{edstedt2024dedode} further support this separation in their recent  ``Detect, Don’t Describe'' framework, summarized as DeDoDe. In DeDoDe, the detector is trained independently to identify repeatable, 3D-consistent keypoints, while the descriptor networks are trained at a later stage. Whereas this decoupled design achieves state-of-the-art results on benchmarks such as MegaDepth-1500, the performance gap between the two descriptors proposed implicitly highlights the central role of high-quality descriptors in feature matching and illustrates that performance is not solely determined by keypoint repeatability.
Another example of decoupled design is \mbox{S-TREK}~\cite{santellani2023strek}, a reinforcement learning–inspired feature extractor with a rotation-equivariant detector that excels at detecting repeatable keypoints across views. S-TREK integrates translation- and rotation-equivariant layers into the detector and employs a reinforcement-learning–inspired sequential strategy to maximize repeatability under large in-plane rotations, while maintaining a lightweight descriptor head.
Similarly, XFeat~\cite{potje2024xfeat} targets runtime scalability: it combines a compute-efficient backbone with a dedicated keypoint branch and a semi-dense match-refinement module, achieving state-of-the-art accuracy while running up to $9\times$ faster than competing methods on low-resources device.

Building on these insights, we focus on descriptor robustness and scalability, addressing DeDoDe's descriptors computational demand with a lightweight architecture that yields strong descriptors at lower compute and memory cost. Our model can be trained on top of any keypoint detector, thus enabling flexible integration into existing pipelines. As we demonstrate in the following sections, this design choice leads to competitive or superior results on multiple challenging benchmarks.

\section{Method}\label{sec:method}

For our descriptor model, we adopt a fully convolutional U-Net-like architecture \cite{ronneberger2015unet}, as it provides a good trade-off between local detail preservation and computational efficiency. 
The proposed architecture begins with an initial $K \times K$ convolution projecting the input \( I \in \mathbb{R}^{3 \times H \times W} \) into a higher-dimensional feature space. This initial step is followed by four consecutive down- and up-sampling blocks, each implemented through Residual U-Net Blocks with Attention~(RUBA). 
The model computes a final L2-normalized descriptor volume \( V \in \mathbb{R}^{128 \times H \times W} \).

\subsection{Residual U-Net Block with Attention}
Each Residual U-Net Block with Attention (RUBA) consists of a main (top) and a residual (bottom) paths, as illustrated in Figure~\ref{fig:res_block}. The main path resizes the input either by downsampling via average pooling or by upsampling via bilinear interpolation, followed by feature concatenation. A $1\times1$ convolution without bias is then used to align the main path with the residual path. The residual path adopts the \emph{pre-activation layer} from He et al.~\cite{he2016identity}, applying batch normalization, activation, and convolution three times in sequence, without weight sharing. The resulting residual features are refined using a Convolutional Block Attention Module (CBAM) \cite{woo2018cbam}.
CBAM applies two lightweight attention mechanisms: the \textit{channel attention} and the \textit{spatial attention}. The former mechanism, denoted as $M_c(\cdot)$, prioritizes the most informative feature channels by recalibrating their activations. The latter mechanism, $M_s(\cdot)$, enhances the most significant spatial regions within each feature map. 
Given a feature map $F \in \mathbb{R}^{C \times H \times W}$, CBAM first applies $M_c(\cdot)$ to enhance relevant channels and then $M_s(\cdot)$ to refine key spatial locations, thus enhancing the overall representational power of the feature map.
The process can be expressed as defined in~\cite{woo2018cbam}: \(F' = M_c(F) \odot F\) and \(F'' = M_s(F') \odot F'\), where \(F'\) and \(F''\) are the intermediate and final output feature maps, respectively, and \(\odot\) denotes element-wise multiplication.

This sequential refinement allows CBAM to capture both channel-wise and spatial dependencies effectively.
For further details on these attention mechanisms, we refer the reader to \cite{woo2018cbam}. The refined residual is then added back to the main path.

All convolutions use kernel size \(K\) and stride \(1\), with zero-padding \(\lfloor K/2 \rfloor\) to preserve spatial resolution. We fix \(K=5\) for convolutions in the residual paths, unless stated otherwise. Within CBAM, the channel-attention module \(M_c(\cdot)\) is a two-layer MLP with reduction ratio \(r=16\) to model inter-channel dependencies, whereas the spatial-attention module \(M_s(\cdot)\) uses a \(7\times7\) convolution (independent of \(K\)) to capture broader spatial context. GELU~\cite{hendrycks2016gelu} is used as the activation function throughout the network, including within CBAM, owing to its smooth nonlinearity and strong empirical performance in deep architectures.

 \begin{figure}[t] 
    \centering
    \includegraphics[width=\linewidth]{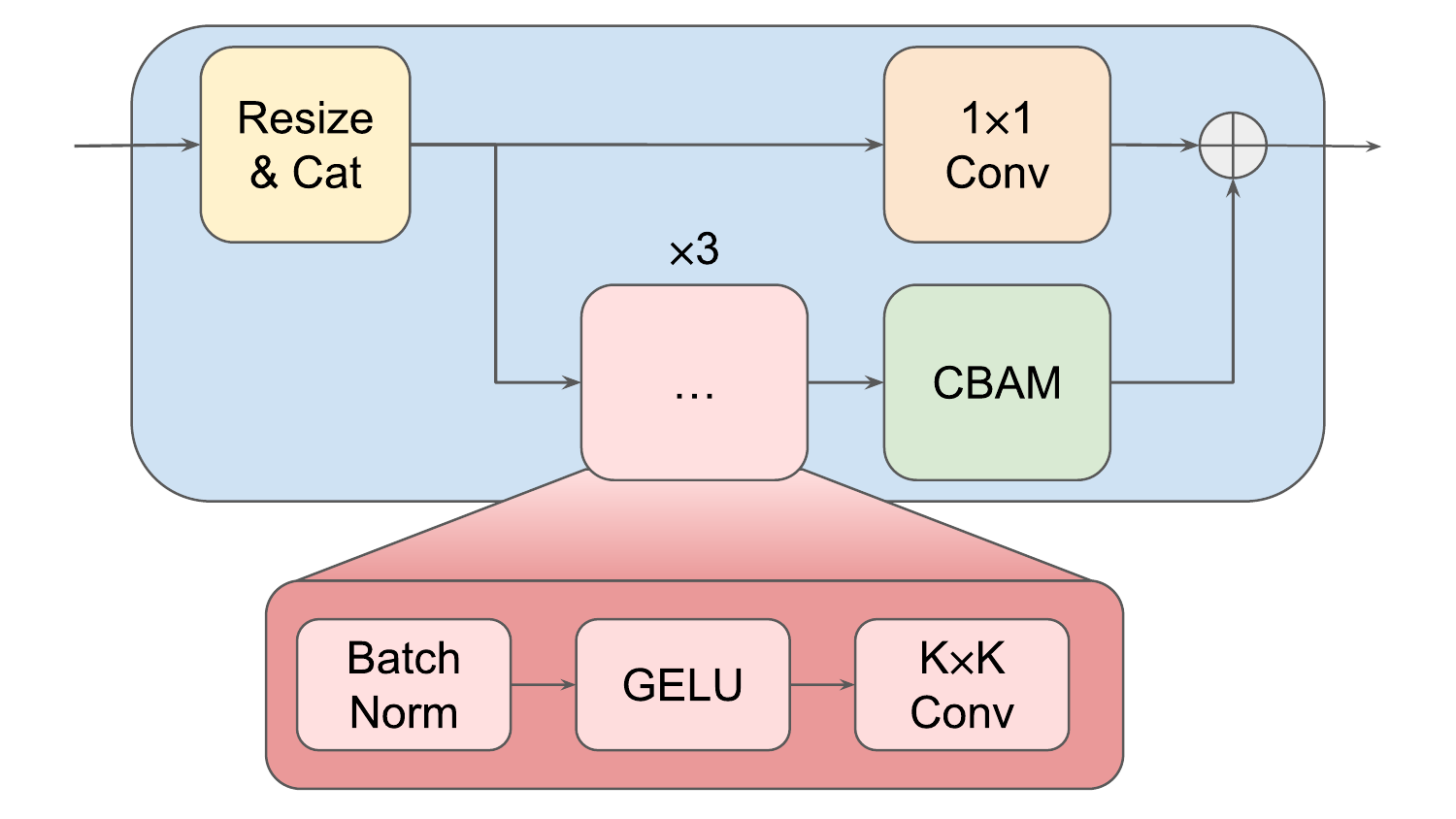}
    \caption{\textbf{Residual U-Net Block with Attention (RUBA).} This block serves as the fundamental building unit of the four-layer U-Net. We use four blocks in the encoder and four in the decoder, configured to resize the features through pooling and upsampling, respectively.}  
    \label{fig:res_block}
    \vspace{-5pt}
\end{figure}

\subsection{Loss}
We employ a variation of the triplet loss~\cite{mishchuk2017working}. 
Local features are first matched using the Mutual Nearest Neighbor (MNN) criterion. 
When a match is established, we form a triplet of descriptors as follows: 
(i) the \textbf{anchor} \(A\), taken from the first image; 
(ii) the \textbf{positive} \(P\), the corresponding matched descriptor in the second image; and 
(iii) the \textbf{negative} \(N\), the descriptor in the second image that achieves the second-best match score with \(A\), selected according to the hardest-negative mining strategy of~\cite{mishchuk2017working}. 

We compute the similarity scores $s$ as:
\begin{equation}
    s_p = A \cdot P, \quad s_n = A \cdot N,
\end{equation}
where \(\cdot\) denotes the dot product. 
Only triplets \(t \in T\) that violate the margin constraint $s_p - s_n < m$, with margin \(m\), are retained. 
This procedure is applied bidirectionally for each image pair (image~1~$\rightarrow$~image~2 and image~2~$\rightarrow$~image~1). 
The final triplet loss is defined as:
\begin{equation}
    \mathcal{L}_{\mathrm{Triplet}}
    = \frac{1}{|T|}
      \sum_{t \in T}
      \bigl( s_{n}^t - s_{p}^t \bigr).
\end{equation}

\noindent To stabilize training and avoid premature plateauing due to overly difficult triplets, we incorporate a curriculum learning strategy: the hardest negative is used with probability \( 1 - \gamma \), and a random negative is chosen otherwise. \( \gamma \) is decayed after each training step, gradually introducing harder examples.

\section{Training Setup}
\label{sec:training}

\subsection{Dataset}
We train our network on the MegaDepth \cite{li2018megadepth} dataset following the protocols described in previous works \cite{tyszkiewicz2020disk, santellani2023strek, edstedt2024dedode, edstedt2024dedode2}. The dataset includes images, camera poses, intrinsic parameters, and depth maps obtained from 3D-reconstructed scenes. We adopt the dataset split proposed by DISK \cite{tyszkiewicz2020disk}, excluding scenes overlapping with the Image Matching Benchmark \cite{jin2021image}. For each scene, we randomly select image pairs from a predefined list of 10\,000 triplets. Each image is resized such that its shortest side measures 512 pixels, then the longer dimension is cropped to produce square images. Additionally, images undergo random rotations with angles uniformly sampled from the range \([-30^\circ, +30^\circ]\).

\subsection{Details}
We follow the approach of training the descriptor on top of prior keypoint detections, as proposed in \cite{santellani2023strek, edstedt2024dedode}. For each method evaluated, we \emph{separately} train a dedicated descriptor network, ensuring that the resulting models are specifically optimized for the corresponding keypoint detectors.

We use the AdamW~\cite{loshchilov2017adamw} optimizer with \(\beta_1 = 0.9\), \(\beta_2 = 0.999\), and a weight decay of 0.01. At the beginning of the training, we linearly increase the learning rate from the minimum value \(\eta_{\text{min}}\) to the maximum value \(\eta_{\text{max}}\) over the first \(w\) warm-up steps to carefully build the gradient moments. After the warm-up phase, the learning rate is decayed exponentially by a factor \(d\) at every iteration, until \(\eta_{\text{min}}\) is reached. 
We set \(\eta_{\text{max}} = 0.005\), \(\eta_{\text{min}} = 0.0001\),  \(w = 2048\), and \(d = 0.99996\).  

For the loss function, we initialize parameter \(\gamma \) to 1 and apply an exponential decay with a factor of 0.9993 at each training step. The loss margin \(\alpha\) is fixed at 0.5.

We adopt automatic mixed-precision training to increase memory efficiency and enable larger batch sizes. Specifically, we use a batch size of 14 on a single NVIDIA RTX 4090 GPU. The model is trained on approximately 250\,000 image pairs. The total training time depends on the speed of the chosen detector.

\section{Experiments}
\label{sec:experiments}

We evaluate our descriptors on several benchmarks, comparing them against the following state-of-the-art methods.
\begin{itemize}
    \item \textbf{SuperPoint} \cite{detone2018superpoint} is an all-in-one model that jointly trains keypoints and descriptors, first on synthetic images for corner detection and then using homographic adaptation for fine-tuning, all within a self-supervised training framework. 
    The authors employ a hinge-loss-based function that minimizes the L2-distance between descriptor pairs at corresponding locations, while maximizing distances for non-corresponding pairs. It is trained on Synthetic Shapes \cite{detone2018superpoint} and MS-COCO 2014 \cite{lin2014microsoft}.
    
    \item \textbf{DISK}~\cite{tyszkiewicz2020disk} is an end-to-end trainable U-Net-based model that learns keypoint detection and description jointly using a policy gradient reinforcement learning framework. It leverages ground-truth geometry (including depth) to assign positive rewards to correct feature matches and vice versa. It is trained on Megadepth \cite{li2018megadepth}.

    
    \item \textbf{RIPE} \cite{kunzel2025ripe} is a reinforcement learning–based framework that trains a VGG-19 model pretrained on ImageNet as an all-in-one detector–descriptor. It maximizes an epipolar geometry–based reward to guide learning, requiring only labels indicating whether two images belong to the same scene. It extracts descriptors from multi-scale hyper-column features. RIPE is trained on the MegaDepth \cite{li2018megadepth}, Tokyo 24/7 \cite{torii2015tokyo}, and ACDC \cite{sakaridis2021acdc} datasets.
    
    \item \textbf{ALIKED}~\cite{zhao2023aliked} builds upon 
    ALIKE~\cite{zhao2022alike}, incorporating a Sparse Deformable Descriptor Head (SDDH) that extracts descriptors exclusively at keypoint locations. Descriptors are trained with a sparse variation of the Neural Reprojection Error loss \cite{zhao2022alike, germain2021neural}. It is trained on Megadepth \cite{li2018megadepth}, R2D2 \cite{revaud2019r2d2}, Oxford and Paris \cite{radenovic2018revisiting}, and the Aachen \cite{sattler2018benchmarking} datasets.
    
    \item \textbf{DeDoDe}~\cite{edstedt2024dedode} separately trains detector and descriptor models starting from the same pre-trained VGG-19 model. The authors propose \emph{two descriptors} denoted as -B and -G, paired with the \emph{same detector}. The -G version incorporates DINOv2 \cite{oquab2024dinov2} features, resulting in a significantly larger model. Both descriptors are trained via a MNN negative log-likelihood loss on Megadepth \cite{li2018megadepth}. 
\end{itemize}
To ensure a fair comparison, we re-evaluated all methods under the same experimental conditions.
We compare them using both their original descriptors and \textup{SAN\textsc{Desc}} across three tasks: \emph{homography estimation} on HPatches, \emph{stereo matching} on the Phototourism test set from the Image Matching Challenge 2021, and finally \emph{stereo pose recovery} on MegaDepth-1500 and on our newly introduced Graz4K dataset.

\subsection{HPatches}
HPatches \cite{balntas2017hpatches} is a widely used benchmark consisting of image sequences exhibiting either viewpoint changes or illumination variations. Following the protocol of \cite{dusmanu2019d2net, santellani2023strek}, we evaluate on a subset of 108 scenes, each comprising one reference image and five target images with corresponding ground-truth homographies. We fix the keypoints budget to 2048 and perform feature matching using MNN. Performance metrics are computed pairwise between the reference and each target. 
We report the \textbf{Mean Matching Accuracy (MMA)} as the percentage of correct matches within a threshold of $\epsilon$ pixels; the \textbf{Matching Score (MS)} as the number of correct matches (up to a pixel threshold) divided by the average number of keypoints in the overlapping area between the two images; and the \textbf{Homography Accuracy (Hom.~Acc.)} as the Area Under the Curve (AUC) of the percentage of estimated homographies with corner error below $\epsilon$ \cite{detone2018superpoint, santellani2023strek}. The corner error is computed as the average distance between the four corners of the reference image and the corresponding corners of the target image warped using the estimated homography. Following \cite{santellani2023strek}, we compute the relative homography using the OpenCV \texttt{findHomography} function with multiple RANSAC thresholds and report the highest Hom.~Acc. score for each method. 
This last metric is often considered the most important for this benchmark, as it measures the effectiveness on a typical downstream SfM task.

We do not include the repeatability metric in our experiments, neither here nor later, as it relates only to the keypoints, which are not modified by \textup{SAN\textsc{Desc}}.

\begin{table}
    \centering
    \resizebox{\linewidth}{!}{%
        \begin{tabular}{lccc|ccc|ccc}
            \toprule
             & \multicolumn{3}{c|}{\textbf{MMA}$\uparrow$} & \multicolumn{3}{c|}{\textbf{MS}$\uparrow$} & \multicolumn{3}{c}{\textbf{Hom. Acc.}$\uparrow$} \\
             
            \textbf{Method} & @1px & @2px & @3px & @1px & @2px & @3px & @1px & @2px & @3px \\
            \midrule
            SuperPoint & 30.3 & 50.6 & 62.2 & 19.0 & 31.5 & 38.4 & 37.6 & 63.7 & 75.0 \\
            \rotatebox[origin=c]{180}{$\Lsh$} w/ \textup{SAN\textsc{Desc}} & \textbf{31.7} & \textbf{53.7} & \textbf{65.8} & \textbf{20.9} & \textbf{35.0} & \textbf{42.6} & \textbf{41.1} & \textbf{66.3} & \textbf{78.3} \\
            \midrule
            DISK & \textbf{45.4} & \textbf{67.1} & \textbf{76.3} & 27.7 & 39.9 & 44.8 & 39.8 & 61.9 & 71.3 \\
            \rotatebox[origin=c]{180}{$\Lsh$} w/ \textup{SAN\textsc{Desc}} & 40.9 & 59.9 & 68.1 & \textbf{27.8} & \textbf{40.1} & \textbf{45.3} & \textbf{42.0} & \textbf{64.8} & \textbf{75.7} \\


            \midrule
            RIPE & \textbf{38.3} & \textbf{58.2} & 67.3 & 25.4 & 38.0 & 43.6 & 44.2 & 67.8 & \textbf{78.9} \\
            \rotatebox[origin=c]{180}{$\Lsh$} w/ \textup{SAN\textsc{Desc}} & 38.0 & \textbf{58.2} & \textbf{67.8} & \textbf{25.9} & \textbf{39.1} & \textbf{45.2} & \textbf{44.3} & \textbf{70.6} & 78.5 \\

            \midrule
            ALIKED & \textbf{41.8} & \textbf{64.3} & \textbf{73.6} & \textbf{28.6} & \textbf{43.0} & \textbf{48.8} & 40.4 & 66.7 & \textbf{77.6} \\
            \rotatebox[origin=c]{180}{$\Lsh$} w/ \textup{SAN\textsc{Desc}} & 41.1 & 62.5 & 70.9 & \textbf{28.6} & 42.6 & 48.0 & \textbf{42.6} & \textbf{68.1} & 77.0 \\

            \midrule
            DeDoDe-B & 44.7 & 62.0 & 69.2 & 22.4 & 30.6 & 33.9 & 50.6 & 71.3 & 79.3 \\
            DeDoDe-G & \textbf{45.6} & \textbf{63.9} & \textbf{71.7} & \textbf{22.7} & \textbf{31.3} & \textbf{34.8} & 50.4 & 71.3 & 78.7 \\
            \rotatebox[origin=c]{180}{$\Lsh$} w/ \textup{SAN\textsc{Desc}} & 44.2 & 60.5 & 67.0 & 22.4 & 30.2 & 33.2 & \textbf{52.0} & \textbf{71.5} & \textbf{79.4} \\

            \bottomrule
        \end{tabular}
    }
    \caption{HPatches results with budget of 2048 keypoints. DeDoDe-B and -G share the same detector. DeDoDe+\textup{SAN\textsc{Desc}} overall achieves the highest Hom.~Acc.}
    \label{tab:hpatches}
\end{table}
As shown in Table~\ref{tab:hpatches}, \textup{SAN\textsc{Desc}} trades a small drop in MMA for higher geometric reliability, with MS  preserved or improved and Hom.~Acc. frequently increasing. The gains are pronounced for SuperPoint, where Hom.~Acc. rises by 3.5 at 1 px and MS improves by up to 4.2. With DISK, MS improves slightly and Hom.~Acc. increases consistently. RIPE and ALIKED show mixed changes in MMA yet still gain in Hom.~Acc. at tight thresholds. Overall, paired with DeDoDe, \textup{SAN\textsc{Desc}} achieves the best Hom.~Acc. across all methods: 52.0 at 1 px, 71.5 at 2 px, and 79.4 at 3 px.

\subsection{The Image Matching Challenge 2021}
The Image Matching Challenge 2021 (IMC) \cite{jin2021image} evaluates local feature matching methods in complex real-world settings. We adopt the IMC 2021 Phototourism test set, which consists of nine scenes, each comprising 100 tourist photos captured with diverse cameras, viewpoints, and lighting conditions. Each scene is identified by its capitals. Images within each scene are exhaustively compared. Following the authors’ protocol, we evaluate estimated poses using the AUC of relative pose accuracy with a 5 degrees error threshold. Pose error is defined as the larger of the rotation and translation errors, with failures assigned when this error exceeds 10 degrees. For each method, we first extract keypoints and descriptors with its original pipeline, then replace the descriptors with \textup{SAN\textsc{Desc}} for a fair comparison. The results are summarized in Table \ref{tab:IMB}.

\begin{table}[t]
    \resizebox{0.475\textwidth}{!}{%
        \begin{tabular}{l|*{9}{c}>{\columncolor{Orchid!50}}c}
            \toprule
            \textbf{Method} & BM & FLC & LM & LB & MC & MR & PSM & SF & SPC & \textbf{AVG} \\
            \midrule
            SuperPoint & 14.1 & 33.6 & 37.9 & 30.4 & 13.7 & 13.0 & 7.1 & 24.9 & 17.6 & 21.3 \\
            \rotatebox[origin=c]{180}{$\Lsh$} w/ \textup{SAN\textsc{Desc}} & \textbf{25.6} & \textbf{50.0} & \textbf{40.5} & \textbf{45.9} & \textbf{23.9} & \textbf{17.9} & \textbf{10.2} & \textbf{36.1} & \textbf{34.8} & \textbf{31.6} \\
            \midrule
            DISK & \textbf{24.1} & 54.7 & \textbf{52.4} & \textbf{56.5} & 35.4 & 26.5 & \textbf{13.7} & \textbf{44.3} & 42.3 & \textbf{38.8} \\
            \rotatebox[origin=c]{180}{$\Lsh$} w/ \textup{SAN\textsc{Desc}} & 24.0 & \textbf{56.0} & 50.6 & 54.7 & \textbf{36.5} & \textbf{26.8} & 12.7 & 44.0 & \textbf{44.0} & \textbf{38.8} \\
            \midrule
            

            RIPE & 18.7 & 53.6 & \textbf{48.9} & \textbf{54.0} & 27.4 & 19.9 & 10.2 & 39.1 & 41.1 & 34.7 \\
            \rotatebox[origin=c]{180}{$\Lsh$} w/ \textup{SAN\textsc{Desc}} & \textbf{25.5} & \textbf{55.0} & 44.5 & 53.3 & \textbf{31.8} & \textbf{25.1} & \textbf{12.3} & \textbf{42.1} & \textbf{44.0} & \textbf{37.0} \\

            \midrule
            ALIKED & \textbf{27.7} & 61.0 & \textbf{44.2} & 54.2 & 37.7 & 25.1 & \textbf{17.2} & \textbf{46.5} & 47.9 & 40.1 \\
            \rotatebox[origin=c]{180}{$\Lsh$} w/ \textup{SAN\textsc{Desc}} & 19.5 & \textbf{61.2} & 41.7 & \textbf{58.8} & \textbf{41.9} & \textbf{30.9} & 15.6 & 46.3 & \textbf{50.7} & \textbf{40.7} \\
            \midrule
            DeDoDe‑B & \underline{26.8} & 52.6 & \underline{34.5} & \underline{45.2} & \underline{32.4} & \underline{19.9} & \underline{10.8} & 37.7 & \underline{44.4} & 33.8 \\
            DeDoDe‑G & \textbf{28.3} & \textbf{55.2} & \textbf{36.8} & \textbf{56.4} & \textbf{34.3} & 19.1 & \textbf{12.8} & \textbf{43.4} & \textbf{46.6} & \textbf{36.9} \\
            \rotatebox[origin=c]{180}{$\Lsh$} w/ \textup{SAN\textsc{Desc}} & 24.7 & \underline{53.4} & 34.2 & 45.1 & 30.3 & \textbf{24.7} & 10.4 & \underline{41.2} & 44.2 & \underline{34.2} \\
            \bottomrule
        \end{tabular}%
    }

    \caption{IMC21 results in terms of AUC@5. The keypoints budget is set to 2048. DeDoDe-B and -G share the same detector. ALIKED+\textup{SAN\textsc{Desc}} overall achieves the highest average score. 
    }
    \label{tab:IMB}
    \vspace{-5pt}
\end{table}

The evaluation shows that combining \mbox{ALIKED} with \textup{SAN\textsc{Desc}} attains the best average performance (AVG) on this benchmark, demonstrating the effectiveness of our descriptor. Overall, all methods except DeDoDe-G benefit from pairing with \textup{SAN\textsc{Desc}}.  In particular, \mbox{SuperPoint} and \mbox{RIPE} exhibit substantial improvements with \textup{SAN\textsc{Desc}}, whereas \mbox{ALIKED}, \mbox{DISK}, and \mbox{DeDoDe-B} achieve results comparable to their native descriptors.

\subsection{Megadepth-1500}
MegaDepth-1500 (MD1500), originally introduced in LoFTR~\cite{sun2021loftr}, is a curated subset of the MegaDepth dataset and has since been adopted in several follow-up works~\cite{edstedt2024dedode, edstedt2024dedode2, lindenberger2023lightglue, edstedt2024roma, gleize2023silk}. The pairs are chosen to maintain a uniform covisibility ratio across the dataset, in contrast to the IMC, where the covisibility distribution follows a Gaussian-like shape \cite{jin2021image}. We use the same evaluation protocol used by \cite{edstedt2024dedode, edstedt2024dedode2} and a score of 90 degrees is given when the fundamental matrix recovery fails. 
Table~\ref{tab:MD1500} presents the results for all the evaluated methods, both with original and \textup{SAN\textsc{Desc}} descriptors, at two keypoints budgets set to 2048 and 30\,000. 

\begin{table}
    \centering
    \resizebox{\linewidth}{!}{%
        \begin{tabular}{l@{}cc|cc}
            \toprule
            \textbf{Keypoints budget$\rightarrow$} & \multicolumn{2}{c}{2048} & \multicolumn{2}{c}{30\,000} \\
            \textbf{Method} & \textbf{AUC@5}$\uparrow$ & \textbf{AUC@10}$\uparrow$ & \textbf{AUC@5}$\uparrow$ & \textbf{AUC@10}$\uparrow$ \\
            \midrule
            SuperPoint & 30.1 & 44.6 & 15.6 & 29.2 \\
            \rotatebox[origin=c]{180}{$\Lsh$} w/ \textup{SAN\textsc{Desc}} & \textbf{42.0} & \textbf{59.2} & \textbf{35.1} & \textbf{52.4} \\
            \midrule
            DISK & 35.3 & 52.0 & 43.2 & 58.1 \\
            \rotatebox[origin=c]{180}{$\Lsh$} w/ \textup{SAN\textsc{Desc}} & \textbf{37.0} & \textbf{54.5} & \textbf{46.0} & \textbf{61.4} \\
            \midrule
            
            RIPE & 43.1	& 59.1 & 36.3 & 52.3 \\
            \rotatebox[origin=c]{180}{$\Lsh$} w/ \textup{SAN\textsc{Desc}} & \textbf{43.6} & \textbf{59.4} & \textbf{42.1} & \textbf{57.7} \\
            \midrule
            ALIKED & 42.0 & 57.5 & 41.5 & 56.5 \\
            \rotatebox[origin=c]{180}{$\Lsh$} w/ \textup{SAN\textsc{Desc}} & \textbf{43.4} & \textbf{60.1} & \textbf{48.3} & \textbf{63.0} \\
            \midrule
            DeDoDe-B & 43.2 & 59.8 & 51.1 & 65.6 \\
            DeDoDe-G & \underline{46.1} & \textbf{63.5} & \textbf{55.3} & \textbf{71.2} \\
            \rotatebox[origin=c]{180}{$\Lsh$} w/ \textup{SAN\textsc{Desc}} & \textbf{46.2} & \underline{62.5} & \underline{52.2} & \underline{67.1} \\
            \bottomrule
        \end{tabular}
    }
    \caption{MD1500 results with 2048 and 30\,000 keypoints budgets. DeDoDe-B and -G share the same detector. DeDoDe+\textup{SAN\textsc{Desc}} scores the highest AUC@5 with 2048 keypoints.}
    \label{tab:MD1500}
    \vspace{-8pt}
\end{table}

At the 5 degrees (2048 keypoints), all detectors benefit from the use of \textup{SAN\textsc{Desc}} descriptors with the combination DeDoDe+\textup{SAN\textsc{Desc}} achieving the highest accuracy. In all the other columns, \textup{SAN\textsc{Desc}} outperforms all the other descriptors except DeDoDe-G.  

Notably, similar to the results in Table~\ref{tab:hpatches}, \textup{SAN\textsc{Desc}} descriptors allow SuperPoint to significantly improve from 30.1 to 42.0 at 2048 keypoints, a relative gain of 40\%, thus outperforming more recent methods like DISK. Interestingly, SuperPoint, RIPE, and ALIKED lose accuracy when the keypoints budget is significantly increased. \textup{SAN\textsc{Desc}} mitigates this issue, enabling stable performance in high-keypoint regimes and delivering substantial gains.

\subsection{Graz4K}\label{sec:ghr} 
High-resolution imagery is often essential for capturing fine structure and achieving high accuracy. However, this regime stresses the compute and memory budgets of feature extractors, exposing scalability limits. To evaluate models under these conditions, we curate a new dataset, called Graz4K, comprising six urban scenes captured with three cameras.
Each camera was calibrated with OpenCV’s ArUco workflow; these intrinsics were then supplied as priors for sparse reconstruction in COLMAP \cite{schoenberger2016sfm} using the default settings. 
Across all scenes, the resulting sparse models achieved a mean reprojection error of $0.97 \pm 0.54$ px over a total of 1\,331\,640 3D points. More details can be found in Supplementary Material. Footage was recorded in 4K at 30 fps and sampled at 1 fps.
Then, we export the view graphs from COLMAP databases and prune them keeping only every tenth image pair. We further discard pairs that are too easy or too hard, retaining a pair only if it yields between 100 and 1\,000 matches. After filtering, the benchmark comprises 1\,866 images and 4\,413 image pairs. We evaluate at three resolutions: native 4K (3840×2160) and downscaled QHD (2560×1440) and FHD (1920×1080). Results were computed following the MD1500 protocol on an NVIDIA RTX 4090 with 24GB using mixed precision.


As reported in Table \ref{tab:high_res}, \textup{SAN\textsc{Desc}} improves over the original descriptors in all scenes and at each resolution. Notably, as image resolution increases, \textup{SAN\textsc{Desc}} delivers greater performance gains and is less sensitive to these increases, as evidenced by the large drops in the original models. 
Furthermore, in this benchmark, SuperPoint exhibits a remarkable improvement when coupled with \textup{SAN\textsc{Desc}}, especially at 4K, achieving the highest gain of +78\%. 
The combination of ALIKED and \textup{SAN\textsc{Desc}} is the top-performing method in this evaluation.
DeDoDe-B and -G run out of memory (OOM) when dealing with inputs larger than FHD on a 24GB VRAM GPU. Despite DeDoDe's impressive performance, the substantial memory footprint of both its descriptors makes it unsuitable for scenarios characterized by high resolution images or limited computational resources, thus limiting its general applicability.

\begin{table}
    \centering
    \resizebox{0.875\linewidth}{!}{%
        \begin{tabular}{lccc|ccc}
            \toprule
            \textbf{Resolution $\rightarrow$} & \multicolumn{1}{c}{FHD} & \multicolumn{1}{c}{QHD} & \multicolumn{1}{c}{4K} & \multicolumn{1}{c}{FHD} & \multicolumn{1}{c}{QHD} & \multicolumn{1}{c}{4K} \\
            \textbf{Method} & \multicolumn{3}{c}{\textbf{AUC@5}$\uparrow$} & \multicolumn{3}{c}{\textbf{AUC@10}$\uparrow$} \\
            
            \midrule
            SuperPoint                                   & 39.9 & 38.4 & 32.4 & 54.4 & 52.4 & 44.7 \\
            \rotatebox[origin=c]{180}{$\Lsh$} w/ \textup{SAN\textsc{Desc}} & \textbf{63.0} & \textbf{62.2} & \textbf{57.7} & \textbf{74.8} & \textbf{73.9} & \textbf{69.8} \\
            \midrule
            DISK                                         & 40.8 & 38.2 & 33.5 & 53.4 & 50.1 & 44.8 \\
            \rotatebox[origin=c]{180}{$\Lsh$} w/ \textup{SAN\textsc{Desc}} & \textbf{55.5} & \textbf{54.0} & \textbf{48.3} & \textbf{69.1} & \textbf{66.9} & \textbf{60.9} \\
            \midrule
            RIPE                                         & 54.5 & 46.5 & 34.0 & 67.0 & 59.0 & 45.1 \\
            \rotatebox[origin=c]{180}{$\Lsh$} w/ \textup{SAN\textsc{Desc}} & \textbf{64.0} & \textbf{62.8} & \textbf{56.7} & \textbf{75.5} & \textbf{74.4} & \textbf{68.2} \\
            \midrule
            ALIKED                                       & 55.0 & 51.7 & 44.4 & 67.7 & 64.9 & 57.7 \\
            \rotatebox[origin=c]{180}{$\Lsh$} w/ \textup{SAN\textsc{Desc}} & \textbf{64.0} & \textbf{64.5} & \textbf{61.8} & \textbf{75.9} & \textbf{75.6} & \textbf{73.4} \\
            \midrule
            DeDoDe\textendash B                          & 56.9 & \multicolumn{2}{c|}{OOM} & 70.3 & \multicolumn{2}{c}{OOM}\\
            DeDoDe\textendash G                          & 52.5 & \multicolumn{2}{c|}{OOM} & 66.5 & \multicolumn{2}{c}{OOM} \\
           \rotatebox[origin=c]{180}{$\Lsh$} w/ \textup{SAN\textsc{Desc}} & \textbf{57.4} & \textbf{54.4} & \textbf{52.2} & \textbf{70.7} & \textbf{67.4} & \textbf{65.3} \\
            \bottomrule
        \end{tabular}
    }
    \caption{Graz4K results with a 2048 keypoints budget for three resolutions. DeDoDe-B and -G share the detector. ALIKED+\textup{SAN\textsc{Desc}} achieves the highest scores in all cases.}
    \label{tab:high_res}
    \vspace{-5pt} 
\end{table}

\subsection{Speed Comparison}
Table~\ref{tab:speed} compares per image processing time in milliseconds for keypoint detection and description under a budget of 2048 keypoints, and reports the model size, in millions of parameters, in the corresponding column. All images were processed in FHD on an NVIDIA RTX 4090 with 24GB. 

Overall, methods such as \mbox{DISK}, \mbox{SuperPoint}, \mbox{ALIKED}, and \mbox{RIPE} run slower when paired with \textup{SAN\textsc{Desc}}, since all-in-one pipelines reuse intermediate features to compute descriptors, whereas \textup{SAN\textsc{Desc}} operates directly on raw images. 
By contrast, with the decoupled \mbox{DeDoDe} methods, \textup{SAN\textsc{Desc}} remains competitive: its accuracy matches \mbox{DeDoDe-G} and exceeds \mbox{DeDoDe-B}, while its runtime is close to \mbox{DeDoDe-B} and faster than \mbox{DeDoDe-G}.
\textup{SAN\textsc{Desc}} alone requires approximately 87 ms on our hardware.

\begin{table}
    \centering
    \resizebox{\linewidth}{!}{%
        \begin{tabular}{l|r|rr|rr}
            \toprule
            \multicolumn{1}{l}{} &
            \multicolumn{1}{r}{} &
            \multicolumn{2}{c}{Speed (ms) $\downarrow$} &
            \multicolumn{2}{c}{VRAM (GB) $\downarrow$} \\
            
            \textbf{Method} & \textbf{Size} & \multicolumn{1}{c}{\textbf{Orig.}} & \multicolumn{1}{c}{\textbf{Ours}} &   \multicolumn{1}{c}{\textbf{Orig.}} & \multicolumn{1}{c}{\textbf{Ours}} \\
            \midrule
            DISK       & 0.26  & 62.1$\pm 0.1$ & 135.2$\pm 0.2$ & 6.96 & 7.01 \\ 
            SuperPoint & 0.30  & 18.3$\pm 0.2$ & 114.8$\pm 0.2$ & 3.50 & 7.18 \\
            ALIKED     & 0.32  & 19.3$\pm 0.3$ & 114.2$\pm 0.2$ & 3.89 & 5.76 \\
            RIPE       & 0.24  & 175.9$\pm 0.2$ & 275.0$\pm 0.2$ & 6.55 & 8.42 \\
            DeDoDe-B   & 15.1  & 181.5$\pm 0.1$ & 189.0$\pm 0.2$ & 8.11 & 5.78 \\
            DeDoDe-G   & 323.2 & 316.8$\pm 0.4$ & 189.0$\pm 0.2$ & 9.31 & 5.78 \\
            \bottomrule
        \end{tabular}
    }
    \caption{Columns report the number of parameters in millions (M), the speed (ms) and the VRAM usage (GB) for original methods and with our descriptor, respectively. Our method has 2.4M parameters. Tests use FHD images on a NVIDIA RTX 4090.
}
    \label{tab:speed}
    \vspace{-8pt} 
\end{table}

\subsection{Ablation}
In this section we focus on the impact that different architectural choices and training strategies have on \textup{SAN\textsc{Desc}} results and runtime. Specifically, in Table \ref{tab:ablation}, we evaluate \textup{SAN\textsc{Desc}} descriptor with \mbox{DeDoDe} and \mbox{RIPE} detectors on the MD1500 benchmark. We report AUC@5 and our descriptor inference time in milliseconds while incrementally adding the architectural components and training strategies under examination.

For both detectors, the largest improvement comes from the \textit{random negative decay} strategy, which increases performance by 7.4 and 7.3 points, respectively. This approach stabilizes training by gradually shifting from easier to harder negative samples. Adding residual paths to the RUBA blocks yields further gains of 1.6 and 1.4 points by improving gradient flow during back-propagation. Finally, incorporating the CBAM attention mechanism provides an additional boost of 1.4 and 0.3 points.

\begin{table}[h]
    \centering
    \resizebox{\linewidth}{!}{%
        \begin{tabular}{lcc|c}
            \toprule
            \multicolumn{1}{l}{\textbf{Method}} 
              & \multicolumn{1}{c}{\textbf{DeDoDe}} 
              & \multicolumn{1}{c}{\textbf{RIPE}} 
              & \multicolumn{1}{r}{\textbf{ms}} \\      
            \midrule
            Plain U-Net              & 35.8  & 34.6                      & 30.1$\pm 0.1$ \\
            + Random Negative Decay  & 43.2  & 41.9                    & 30.1$\pm 0.1$\\
            + Residual Paths         & 44.8  & 43.3                      & 37.1$\pm 0.1$ \\
            + Attention (Full)       & \textbf{46.2}  & \textbf{43.6}  & 86.7$\pm 0.1$\\
    
            \bottomrule
        \end{tabular}
    }
    \caption{Ablation study on MD1500 in terms of AUC@5 with 2,048 keypoints. 
    }
    \label{tab:ablation}
\end{table}

\vspace{-5pt} 
\subsection{The Need To Retrain on Each Detector}
\label{sec:discussion}

The training framework we propose in this paper trains a \textbf{\textup{SAN\textsc{Desc}}} model \emph{separately} for each specific detector. Nevertheless, it is possible to train a \textup{SAN\textsc{Desc}} variant on randomly generated keypoints to remain \emph{detector-agnostic} (DA). Specifically, projecting the image grid back and forth between the two images, retaining only points whose reprojection error is below a small pixel threshold, and then sampling uniformly. We call this version \textbf{\textup{SAN\textsc{Desc}} DA}.

To asses descriptors compatibility and the sensitivity of descriptors to detector shift, we run a full cross detector-descriptor evaluation with the methods listed in Section \ref{sec:experiments}. Specifically, we pair each detector with every descriptor under a fixed matching pipeline and test on the MD1500 dataset.

Table~\ref{tab:comparison}'s last column reports the average score across detectors for each descriptor. DeDoDe-G emerges as the most flexible and effective descriptor, thanks to its higher parameter count and extensive training. Nevertheless, both \textup{SAN\textsc{Desc}} and \textup{SAN\textsc{Desc}} DA follow closely falling short by only 0.9 (2\%) and 1.4 (3.2\%) AUC@5 points, respectively. The detector-specific training scheme consistently yields superior performance compared with the DA model, thereby justifying our methodological choice.

\begingroup
\setlength{\tabcolsep}{3pt} 
\begin{table}[t]
  \centering
  \begin{adjustbox}{max width=\linewidth} 
  \begin{tabular}{@{}l|ccccc|c@{}}
    \toprule
     & \multicolumn{5}{c}{\textbf{Detectors}}  \\
     \textbf{Descriptors} & ALIKED & DeDoDe & DISK & RIPE & SuperPoint & AVG \\
    \midrule
    ALIKED      & \fbox{42.0} & 38.5   & 10.3  & 12.7    & 37.8   & 28.3 \\
    DeDoDe-B    & 42.0 & \fbox{43.2} & 35.7 & 41.7 & 40.7 & 40.7 \\
    DeDoDe-G    & \textbf{45.3}   & \fbox{\underline{46.1}}  & \underline{36.9} & \textbf{44.5}  & \textbf{43.9} & \textbf{43.3} \\
    DISK        & 38.5 & 38.3 & \fbox{35.3} & 37.6 & 35.1 & 37.0 \\
    RIPE        & 42.5 & 39.0 & 34.6 & \fbox{43.1} & 37.1 & 39.3  \\
    SuperPoint  & 33.3 & 29.3 & 26.7 & 30.8 & \fbox{30.1} & 30.0 \\
    \midrule
    \textup{SAN\textsc{Desc}} DA   & 41.7 & 44.6 & 36.7 & 42.9 & 41.9 & 41.9 \\
    \textup{SAN\textsc{Desc}}     & \underline{43.4} & \textbf{46.2} &  \textbf{37.0} & \underline{43.6} & \underline{42.0} & \underline{42.4} \\
    \bottomrule
  \end{tabular}
  \end{adjustbox}
  \caption{Cross detector-descriptor evaluation on MD1500 in terms of AUC@5 with 2,048 keypoints budget. Boxed entries highlight native detector–descriptor pairs. Last column reports per-descriptor average score.
  }
  \label{tab:comparison}
  \vspace{-8pt} 
\end{table}
\endgroup

\section{Conclusion}
\label{sec:conclusion}

In this work, we introduced \textbf{\textup{SAN\textsc{Desc}}}, an efficient and robust descriptor architecture that can be trained on top of existing keypoint detectors to improve over their original descriptors. 
\textup{SAN\textsc{Desc}} employs a lightweight U-Net architecture that leverages channel and spatial attention through the proposed \textbf{RUBA} block to produce more robust descriptors. We trained \textup{SAN\textsc{Desc}} using a modified triplet loss combined with a curriculum-learning-inspired hard-negative-mining strategy to stabilize the training.
 
We demonstrated that \textbf{\textup{SAN\textsc{Desc}}} improves multiple existing detectors on several popular benchmarks, including HPatches, the Image Matching Challenge 2021, and MegaDepth-1500. In addition, we introduced the Graz4K datasets showing that \textup{SAN\textsc{Desc}} performance gracefully scales with high-resolution images, which is crucial where accuracy is critical. We further show that our framework remains competitive with the very large DeDoDe-G model, while requiring only a fraction of its cost.

\paragraph{Acknowledgements} This work has been supported by the FFG under Contract No. 881844 within the project ``Pro²Future''.

{
    \small
    \bibliographystyle{ieeenat_fullname}
    \bibliography{main}

@inproceedings{sift,
  author       = {David~G. Lowe},
  title        = {Object Recognition from Local Scale{\textit{-}}Invariant Features},
  booktitle    = {Proceedings of the Seventh IEEE International Conference on Computer Vision (ICCV)},
  volume       = {2},
  pages        = {1150--1157},
  year         = {1999},
  organization = {IEEE},
}

@inproceedings{tian2017l2net,
  author       = {Yurun Tian and Bin Fan and Fuchao Wu},
  title        = {{L2{\textendash}Net}: Deep Learning of Discriminative Patch Descriptor in Euclidean Space},
  booktitle    = {Proceedings of the IEEE Conference on Computer Vision and Pattern Recognition (CVPR)},
  pages        = {661--669},
  year         = {2017},
  organization = {IEEE},
}

@article{mishchuk2017working,
  author  = {Anastasiia Mishchuk and Dmytro Mishkin and Filip Radenovic and Jiri Matas},
  title   = {Working Hard to Know Your Neighbor's Margins: Local Descriptor Learning Loss},
  journal = {Advances in Neural Information Processing Systems},
  volume  = {30},
  pages   = {640--651},
  year    = {2017},
}

@inproceedings{detone2018superpoint,
  author       = {Daniel DeTone and Tomasz Malisiewicz and Andrew Rabinovich},
  title        = {SuperPoint: Self{\textendash}Supervised Interest Point Detection and Description},
  booktitle    = {Proceedings of the IEEE Conference on Computer Vision and Pattern Recognition Workshops (CVPR W)},
  pages        = {224--236},
  year         = {2018},
  organization = {IEEE},
}

@inproceedings{edstedt2024dedode,
  author       = {Johan Edstedt and Georg B{\"o}kman and M{\aa}rten Wadenb{\"a}ck and Michael Felsberg},
  title        = {{DeDoDe}: Detect, Don't Describe—Describe, Don't Detect for Local Feature Matching},
  booktitle    = {Proceedings of the 2024 International Conference on 3D Vision (3DV)},
  pages        = {148--157},
  year         = {2024},
  organization = {IEEE},
}

@inproceedings{li2022decoupling,
  author       = {Kunhong Li and Longguang Wang and Li Liu and Qing Ran and Kai Xu and Yulan Guo},
  title        = {Decoupling Makes Weakly Supervised Local Feature Better},
  booktitle    = {Proceedings of the IEEE/CVF Conference on Computer Vision and Pattern Recognition (CVPR)},
  pages        = {15838--15848},
  year         = {2022},
  organization = {IEEE},
}

@inproceedings{woo2018cbam,
  author       = {Sanghyun Woo and Jongchan Park and Joon{\textendash}Young Lee and In~So Kweon},
  title        = {{CBAM}: Convolutional Block Attention Module},
  booktitle    = {Proceedings of the European Conference on Computer Vision (ECCV)},
  pages        = {3--19},
  year         = {2018},
  organization = {Springer},
}

@article{bay2008surf,
  author = {Bay, Herbert and Ess, Andreas and Tuytelaars, Tinne and Gool, Luc Van},
  title = {Speeded-Up Robust Features (SURF)},
  journal = {Computer Vision and Image Understanding},
  number = 3,
  pages = {346 - 359},
  volume = 110,
  year = 2008
}

@inproceedings{han2015matchnet,
  author       = {Xufeng Han and Thomas Leung and Yangqing Jia and Rahul Sukthankar and Alexander~C. Berg},
  title        = {MatchNet: Unifying Feature and Metric Learning for Patch{\textendash}Based Matching},
  booktitle    = {Proceedings of the IEEE Conference on Computer Vision and Pattern Recognition (CVPR)},
  pages        = {3279--3286},
  year         = {2015},
  organization = {IEEE},
}

@inproceedings{simo2015dldcf,
  author       = {Edgar Simo{\textendash}Serra and Eduard Trulls and Luis Ferraz and Iasonas Kokkinos and Pascal Fua and Francesc Moreno{\textendash}Noguer},
  title        = {Discriminative Learning of Deep Convolutional Feature Point Descriptors},
  booktitle    = {Proceedings of the IEEE International Conference on Computer Vision (ICCV)},
  pages        = {118--126},
  year         = {2015},
  organization = {IEEE},
}

@inproceedings{balntas2016triplet,
  author       = {Vassileios Balntas and Edgar Riba and Daniel Ponsa and Krystian Mikolajczyk},
  title        = {Learning Local Feature Descriptors with Triplets and Shallow Convolutional Neural Networks},
  booktitle    = {Proceedings of the British Machine Vision Conference (BMVC)},
  volume       = {1},
  pages        = {3},
  year         = {2016},
  organization = {BMVA},
}

@inproceedings{tian2019sosnet,
  author       = {Yurun Tian and Xin Yu and Bin Fan and Fuchao Wu and Huub Heijnen and Vassileios Balntas},
  title        = {{SOSNet}: Second Order Similarity Regularization for Local Descriptor Learning},
  booktitle    = {Proceedings of the IEEE/CVF Conference on Computer Vision and Pattern Recognition (CVPR)},
  pages        = {11016--11025},
  year         = {2019},
  organization = {IEEE},
}

@inproceedings{balntas2017hpatches,
  author       = {Vassileios Balntas and Karel Lenc and Andrea Vedaldi and Krystian Mikolajczyk},
  title        = {{HPatches}: A Benchmark and Evaluation of Handcrafted and Learned Local Descriptors},
  booktitle    = {Proceedings of the IEEE Conference on Computer Vision and Pattern Recognition (CVPR)},
  pages        = {5173--5182},
  year         = {2017},
  organization = {IEEE},
}

@article{jin2021image,
  author  = {Yuhe Jin and Dmytro Mishkin and Anastasiia Mishchuk and Jiri Matas and Pascal Fua and Kwang~Moo Yi and Eduard Trulls},
  title   = {Image Matching across Wide Baselines: From Paper to Practice},
  journal = {International Journal of Computer Vision},
  volume  = {129},
  number  = {2},
  pages   = {517--547},
  year    = {2021},
  publisher = {Springer},
}

@inproceedings{dusmanu2019d2net,
  author       = {Mihai Dusmanu and Ignacio Rocco and Tomáš Pajdla and Marc Pollefeys and Josef Sivic and Akihiko Torii and Torsten Sattler},
  title        = {{D2{\textendash}Net}: A Trainable CNN for Joint Detection and Description of Local Features},
  booktitle    = {Proceedings of the IEEE/CVF Conference on Computer Vision and Pattern Recognition (CVPR)},
  pages        = {8092--8101},
  year         = {2019},
  organization = {IEEE},
}

@article{revaud2019r2d2,
  author  = {Jerome Revaud and Cesar De Souza and Martin Humenberger and Philippe Weinzaepfel},
  title   = {{R2D2}: Reliable and Repeatable Detector and Descriptor},
  journal = {Advances in Neural Information Processing Systems},
  volume  = {32},
  pages   = {124--135},
  year    = {2019},
}

@inproceedings{santellani2022mdnet,
  author       = {Emanuele Santellani and Christian Sormann and Mattia Rossi and Andreas Kuhn and Friedrich Fraundorfer},
  title        = {{MD\textendash{}Net}: Multi{\textendash}Detector for Local Feature Extraction},
  booktitle    = {Proceedings of the 26th International Conference on Pattern Recognition (ICPR)},
  pages        = {3944--3951},
  year         = {2022},
  organization = {IEEE},
}

@article{tyszkiewicz2020disk,
  author  = {Micha{\l} Tyszkiewicz and Pascal Fua and Eduard Trulls},
  title   = {{DISK}: Learning Local Features with Policy Gradient},
  journal = {Advances in Neural Information Processing Systems},
  volume  = {33},
  pages   = {14254--14265},
  year    = {2020},
}

@inproceedings{santellani2023strek,
  author       = {Emanuele Santellani and Christian Sormann and Mattia Rossi and Andreas Kuhn and Friedrich Fraundorfer},
  title        = {{S{\textendash}Trek}: Sequential Translation and Rotation Equivariant Keypoints for Local Feature Extraction},
  booktitle    = {Proceedings of the IEEE/CVF International Conference on Computer Vision (ICCV)},
  pages        = {9728--9737},
  year         = {2023},
  organization = {IEEE},
}

@inproceedings{ronneberger2015unet,
  author       = {Olaf Ronneberger and Philipp Fischer and Thomas Brox},
  title        = {{U{\textendash}Net}: Convolutional Networks for Biomedical Image Segmentation},
  booktitle    = {Proceedings of the Medical Image Computing and Computer‐Assisted Intervention (MICCAI)},
  pages        = {234--241},
  year         = {2015},
  organization = {Springer},
}

@inproceedings{he2016identity,
  author       = {Kaiming He and Xiangyu Zhang and Shaoqing Ren and Jian Sun},
  title        = {Identity Mappings in Deep Residual Networks},
  booktitle    = {Proceedings of the European Conference on Computer Vision (ECCV)},
  pages        = {630--645},
  year         = {2016},
  organization = {Springer},
}

@article{hendrycks2016gelu,
  author  = {Dan Hendrycks and Kevin Gimpel},
  title   = {Gaussian Error Linear Units ({GELUs})},
  journal = {arXiv preprint arXiv:1606.08415},
  year    = {2016},
}

@inproceedings{li2018megadepth,
  author       = {Zhengqi Li and Noah Snavely},
  title        = {MegaDepth: Learning Single‐View Depth Prediction from Internet Photos},
  booktitle    = {Proceedings of the IEEE Conference on Computer Vision and Pattern Recognition (CVPR)},
  pages        = {2041--2050},
  year         = {2018},
  organization = {IEEE},
}

@inproceedings{edstedt2024dedode2,
  author       = {Johan Edstedt and Georg B{\"o}kman and Zhenjun Zhao},
  title        = {{DeDoDe v2}: Analyzing and Improving the DeDoDe Keypoint Detector},
  booktitle    = {Proceedings of the IEEE/CVF Conference on Computer Vision and Pattern Recognition (CVPR)},
  pages        = {4245--4253},
  year         = {2024},
  organization = {IEEE},
}

@article{zhao2023aliked,
  author  = {Xiaoming Zhao and Xingming Wu and Weihai Chen and Peter C. Y. Chen and Qingsong Xu and Zhengguo Li},
  title   = {{ALIKED}: A Lighter Keypoint and Descriptor Extraction Network via Deformable Transformation},
  journal = {IEEE Transactions on Instrumentation and Measurement},
  volume  = {72},
  pages   = {1--16},
  year    = {2023},
  publisher = {IEEE},
}

@inproceedings{
loshchilov2017adamw,
title={Decoupled Weight Decay Regularization},
author={Ilya Loshchilov and Frank Hutter},
booktitle={International Conference on Learning Representations},
year={2019},
url={https://openreview.net/forum?id=Bkg6RiCqY7},
}

@article{zhao2022alike,
  author  = {Xiaoming Zhao and Xingming Wu and Jinyu Miao and Weihai Chen and Peter C. Y. Chen and Zhengguo Li},
  title   = {ALIKE: Accurate and Lightweight Keypoint Detection and Descriptor Extraction},
  journal = {IEEE Transactions on Multimedia},
  volume  = {25},
  pages   = {3101--3112},
  year    = {2022},
  publisher = {IEEE},
}

@article{oquab2024dinov2,
  title={DINOv2: Learning Robust Visual Features without Supervision},
  author={Oquab Maxime and Darcet Timoth{\'e}e and Moutakanni Th{\'e}o and Vo Huy and Szafraniec, Marc and Khalidov Vasil and Fernandez Pierre and Haziza Daniel and Massa Francisco and El-Nouby Alaaeldin and others},
  journal={Transactions on Machine Learning Research Journal},
  pages={1--31},
  year={2024}
}

@inproceedings{sun2021loftr,
  author       = {Jiaming Sun and Zehong Shen and Yuang Wang and Hujun Bao and Xiaowei Zhou},
  title        = {{LoFTR}: Detector{\textendash}Free Local Feature Matching with Transformers},
  booktitle    = {Proceedings of the IEEE/CVF Conference on Computer Vision and Pattern Recognition (CVPR)},
  pages        = {8922--8931},
  year         = {2021},
  organization = {IEEE},
}

@inproceedings{lindenberger2023lightglue,
  author       = {Philipp Lindenberger and Paul{\textendash}Edouard Sarlin and Marc Pollefeys},
  title        = {LightGlue: Local Feature Matching at Light Speed},
  booktitle    = {Proceedings of the IEEE/CVF International Conference on Computer Vision (ICCV)},
  pages        = {17627--17638},
  year         = {2023},
  organization = {IEEE},
}

@inproceedings{edstedt2024roma,
  author       = {Johan Edstedt and Qiyu Sun and Georg B{\"o}kman and M{\aa}rten Wadenb{\"a}ck and Michael Felsberg},
  title        = {{RoMa}: Robust Dense Feature Matching},
  booktitle    = {Proceedings of the IEEE/CVF Conference on Computer Vision and Pattern Recognition (CVPR)},
  pages        = {19790--19800},
  year         = {2024},
  organization = {IEEE},
}

@inproceedings{gleize2023silk,
  author       = {Pierre Gleize and Weiyao Wang and Matt Feiszli},
  title        = {{SILK}: Simple Learned Keypoints},
  booktitle    = {Proceedings of the IEEE/CVF International Conference on Computer Vision (ICCV)},
  pages        = {22499--22508},
  year         = {2023},
  organization = {IEEE},
}

@inproceedings{lin2014microsoft,
  title={Microsoft coco: Common objects in context},
  author={Lin, Tsung-Yi and Maire, Michael and Belongie, Serge and Hays, James and Perona, Pietro and Ramanan, Deva and Doll{\'a}r, Piotr and Zitnick, C Lawrence},
  booktitle={European conference on computer vision},
  pages={740--755},
  year={2014},
  organization={Springer}
}

@inproceedings{germain2021neural,
  title={Neural reprojection error: Merging feature learning and camera pose estimation},
  author={Germain, Hugo and Lepetit, Vincent and Bourmaud, Guillaume},
  booktitle={Proceedings of the IEEE/CVF Conference on Computer Vision and Pattern Recognition},
  pages={414--423},
  year={2021}
}

@inproceedings{radenovic2018revisiting,
  title={Revisiting oxford and paris: Large-scale image retrieval benchmarking},
  author={Radenovi{\'c}, Filip and Iscen, Ahmet and Tolias, Giorgos and Avrithis, Yannis and Chum, Ond{\v{r}}ej},
  booktitle={Proceedings of the IEEE conference on computer vision and pattern recognition},
  pages={5706--5715},
  year={2018}
}

@inproceedings{schoenberger2016sfm,
    author={Sch\"{o}nberger, Johannes Lutz and Frahm, Jan-Michael},
    title={{Structure-from-Motion Revisited}},
    booktitle={Conference on Computer Vision and Pattern Recognition (CVPR)},
    year={2016}
}

@inproceedings{potje2024xfeat,
  title={Xfeat: Accelerated features for lightweight image matching},
  author={Potje, Guilherme and Cadar, Felipe and Araujo, Andr{\'e} and Martins, Renato and Nascimento, Erickson R},
  booktitle={Proceedings of the IEEE/CVF Conference on Computer Vision and Pattern Recognition},
  pages={2682--2691},
  year={2024}
}

@inproceedings{sakaridis2021acdc, 
  title={ACDC: The adverse conditions dataset with correspondences for semantic driving scene understanding},
  author={Sakaridis, Christos and Dai, Dengxin and Van Gool, Luc},
  booktitle={Proceedings of the IEEE/CVF international conference on computer vision},
  pages={10765--10775},
  year={2021}
}

@inproceedings{torii2015tokyo,
  title={24/7 place recognition by view synthesis},
  author={Torii, Akihiko and Arandjelovic, Relja and Sivic, Josef and Okutomi, Masatoshi and Pajdla, Tomas},
  booktitle={Proceedings of the IEEE conference on computer vision and pattern recognition},
  pages={1808--1817},
  year={2015}
}

@article{zhou2009object,
  title={Object tracking using SIFT features and mean shift},
  author={Zhou, Huiyu and Yuan, Yuan and Shi, Chunmei},
  journal={Computer vision and image understanding},
  volume={113},
  number={3},
  pages={345--352},
  year={2009},
  publisher={Elsevier}
}

@inproceedings{sattler2018benchmarking,
  title={Benchmarking 6dof outdoor visual localization in changing conditions},
  author={Sattler, Torsten and Maddern, Will and Toft, Carl and Torii, Akihiko and Hammarstrand, Lars and Stenborg, Erik and Safari, Daniel and Okutomi, Masatoshi and Pollefeys, Marc and Sivic, Josef and others},
  booktitle={Proceedings of the IEEE conference on computer vision and pattern recognition},
  pages={8601--8610},
  year={2018}
}

@inproceedings{kunzel2025ripe,
  title={RIPE: Reinforcement Learning on Unlabeled Image Pairs for Robust Keypoint Extraction},
  author={K{\"u}nzel, Johannes and Hilsmann, Anna and Eisert, Peter},
  booktitle={Proceedings of the IEEE/CVF International Conference on Computer Vision},
  pages={4868--4877},
  year={2025}
}
}

\end{document}